\documentclass{article}

\usepackage[main, preprint]{neurips_2026}

\usepackage[utf8]{inputenc} %
\usepackage[T1]{fontenc}    %
\usepackage{hyperref}       %
\usepackage{url}            %
\usepackage{booktabs}       %
\usepackage{amsfonts}       %
\usepackage{nicefrac}       %
\usepackage{microtype}      %
\usepackage{xcolor}         %
\usepackage{enumitem}

\usepackage{amsmath}
\usepackage{amssymb}
\usepackage{mathtools}
\usepackage{amsthm}

\usepackage{subcaption}

\usepackage[pdftex]{graphicx}

\usepackage[capitalize,noabbrev]{cleveref}

\usepackage{siunitx}

\usepackage{pifont}
\newcommand{\cmark}{\ding{51}}%
\newcommand{\xmark}{\ding{55}}%

\newcommand{\bu}{\boldsymbol{u}}

\newcommand{\bx}{\boldsymbol{x}}
\newcommand{\by}{\boldsymbol{y}}
\newcommand{\bz}{\boldsymbol{z}}

\newcommand{\bpi}{\boldsymbol{\pi}}

\definecolor{xblue}{HTML}{1F77B4}
\definecolor{xorange}{HTML}{FF7F0E}
\definecolor{xgreen}{HTML}{2CA02C}

\newcommand{\first}[1]{\textcolor{xblue}{#1}}
\newcommand{\second}[1]{\textcolor{xgreen}{#1}}
\newcommand{\third}[1]{\textcolor{xorange}{#1}}

\newcommand{\modelname}{SLiDE}

\newcommand{\ci}[2]{%
    $\begin{matrix}#2\\#1\end{matrix}$

}

\title{SLiDE: Structured Linear Dynamics \\ for Forecasting with Exogenous Inputs}

\author{%
  Sebastian P\"utz \thanks{Equal contribution.} \\
  Karlsruhe Institute of Technology,\\ 
  Helmholtz AI \\
  \And
  Theodore Glavas \footnotemark[1] \\
  McGill University,\\
  ILLS, Mila \\
  \And
  Benjamin Schäfer \\
  Karlsruhe Institute of Technology,\\
  Helmholtz AI \\
  \And
  Boris N. Oreshkin \thanks{This work is not related to the author's position at Amazon.}\\
  Amazon Science \\
  \And
  Mark Coates \\
  McGill University,\\ ILLS, Mila \\
}

\begin{document}

\maketitle

\begin{abstract}
Accurate time series forecasting with exogenous inputs is critical across domains, including energy and retail, yet modern deep learning models often overfit to correlations that do not generalize under shifts in these inputs. 
  We propose \modelname{}, a Koopman-inspired architecture for forecasting with exogenous inputs that imposes structure on latent temporal dynamics. 
  \modelname{} approximates nonlinear system evolution using a learned linear operator in a latent space, combining (i) a history encoder that reconstructs the current latent state from past targets and inputs with (ii) a shared linear rollout driven by future exogenous variables. By enforcing a common linear recurrence across encoding and prediction, \modelname{} aligns how past information is encoded with how future states evolve, reducing parameterization and preventing reliance on spurious correlations in the input history. 
  Empirically, \modelname{} achieves state-of-the-art accuracy on real-world benchmarks, including electricity price forecasting and retail demand, while maintaining a computational footprint comparable to lightweight MLP architectures. \modelname{} is especially effective at generalizing to shifted exogenous inputs, reducing MSE by 20\% on electricity price samples with out-of-range exogenous values and MAE by 30\% in a synthetic transfer setting where the same dynamical system is driven by an unseen exogenous process. Ablations confirm the importance of both the structured history encoder and the linear latent rollout, suggesting that structured latent linear dynamics provide a useful inductive bias for forecasting with exogenous inputs.
\end{abstract}

\begin{figure}[t]
    \centering

    \begin{minipage}[t]{0.43\textwidth}
        \vspace{0pt}
        \begin{subfigure}[t]{\linewidth}
            \vspace{0pt}
            \centering
            \includegraphics[width=\linewidth]{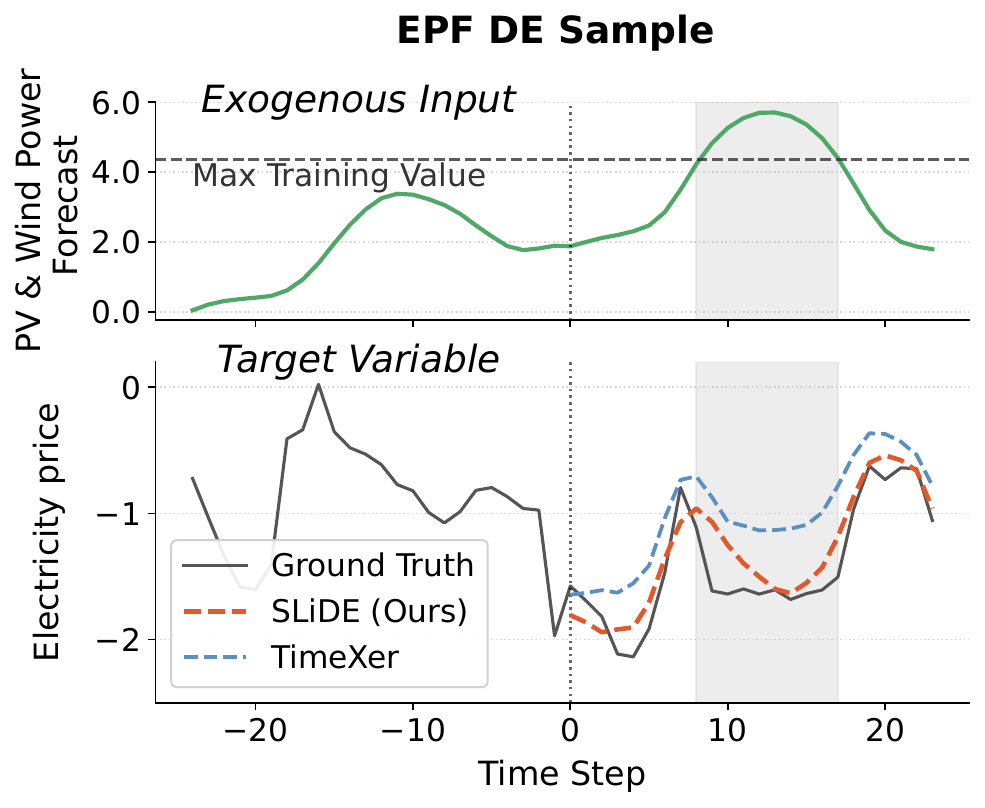}
            
        \end{subfigure}

        \begin{subfigure}[t]{\linewidth}
            \vspace{0pt}
            \centering
            \includegraphics[width=\linewidth]{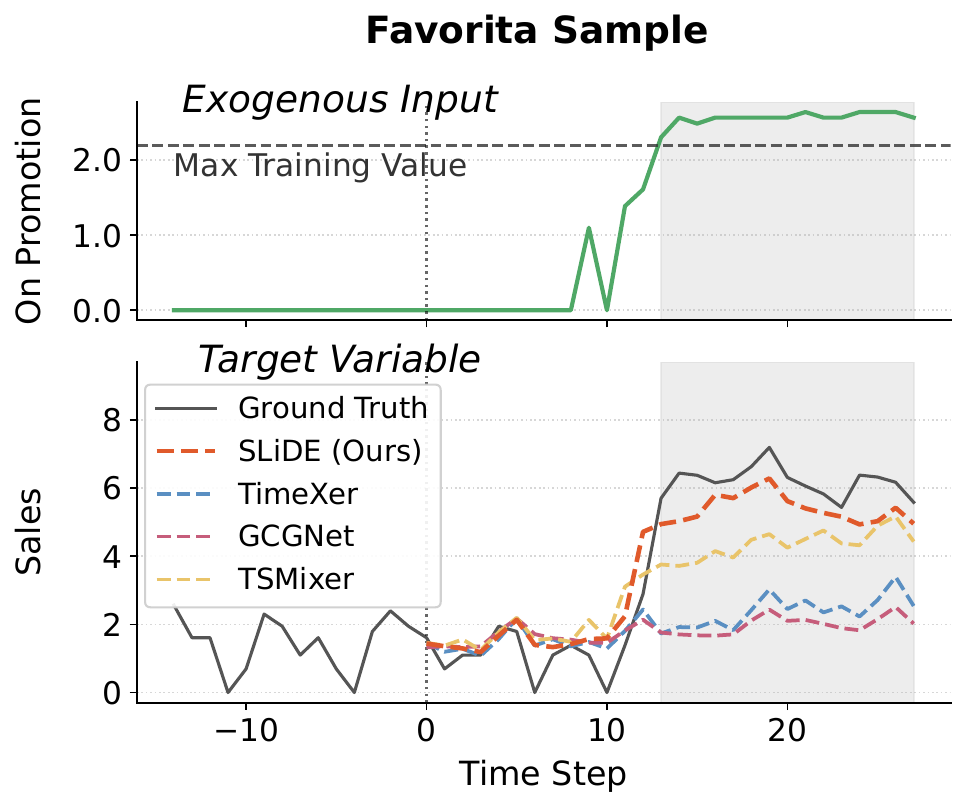}
            \caption{Test samples with out-of-range inputs}
            \label{fig:ood_sample}
        \end{subfigure}
    \end{minipage}
    \hfill
    \begin{minipage}[t]{0.55\textwidth}
        \begin{subfigure}[t]{\linewidth}
            \vspace{0pt}
            \centering
            \includegraphics[width=\linewidth]{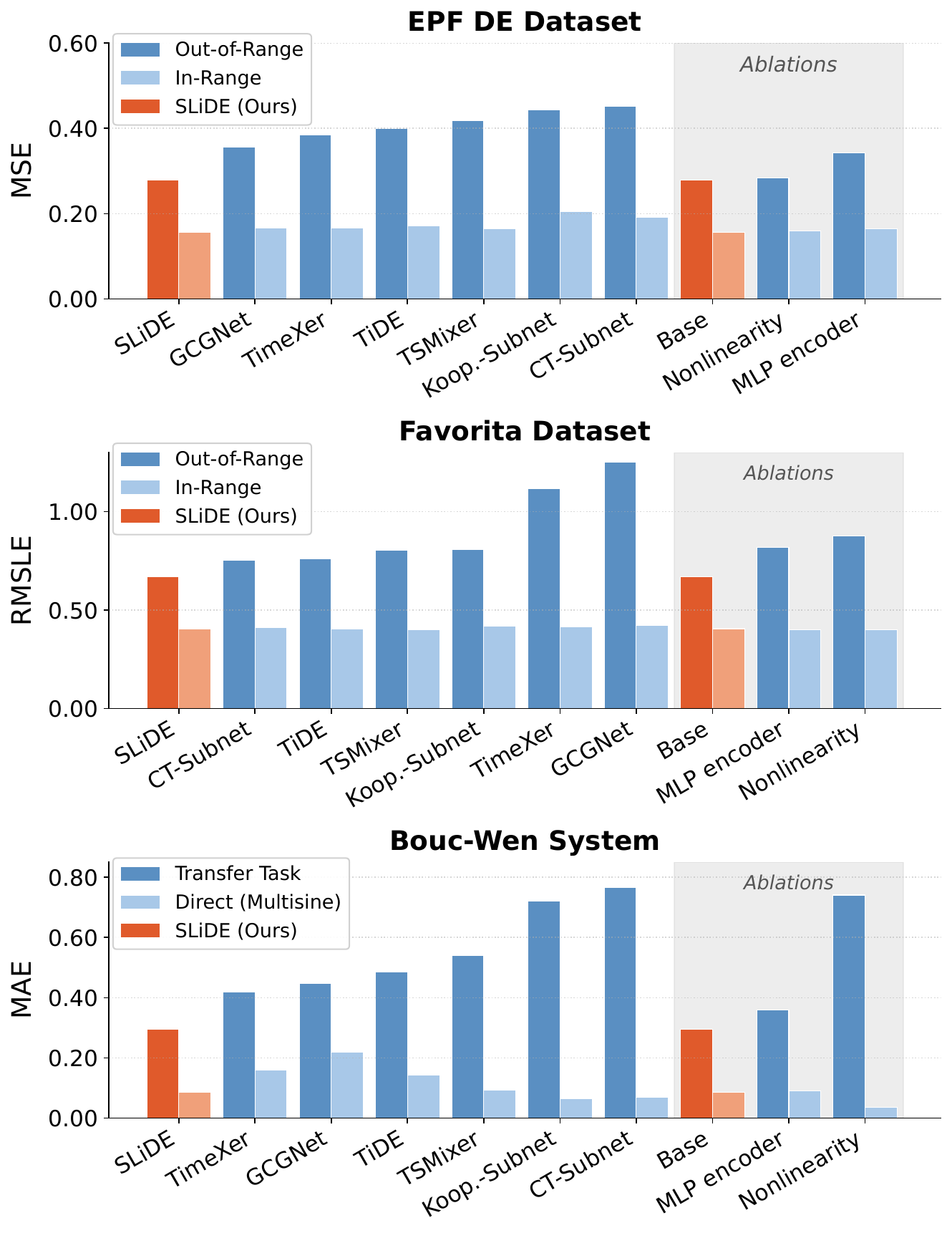}
            \caption{Performance under shifted exogenous inputs}
            \label{fig:ood_barplot}
        \end{subfigure}
    \end{minipage}
    \caption{\modelname{} generalizes well to shifted exogenous inputs, including out-of-range (OOR) input values and unseen test-time input processes. Left: top, an EPF-DE sample in which the wind forecast exceeds the maximum training value, and bottom, a Favorita sample in which the number of related promoted items (log-scale) exceeds the training range. In both cases, \modelname{} stays closer to the ground truth than the baselines. Right: \modelname{} reduces MSE by 20\% on OOR EPF-DE samples and MAE by 30\% on the Bouc-Wen transfer task, where models are tested on an unseen forcing process. Ablations show that the structured encoder and linear rollout are critical for shifted-input performance, as shown by an 18\% lower RMSLE relative to ablations on OOR Favorita samples.}
    \label{fig:ood_showcase}
\end{figure}

\section{Introduction}

Time series forecasting supports decision-making across energy~\citep{hongEnergyForecastingReview2020}, economics~\citep{stock2001forecasting}, healthcare~\citep{harutyunyan2019multitask}, transportation~\citep{vlahogianni2014short}, and retail. In many settings, accurate forecasts depend on exogenous inputs~\citep{box2015time}: electricity prices depend on load and renewable generation forecasts~\citep{weron2014electricity,trebbienUnderstandingElectricityPrices2023a}, while retail demand depends on promotions, holidays, and product- or store-level context~\citep{fildes2022retail}. These inputs are often forecasted, noisy, or generated by changing external processes, making robustness to shifted exogenous inputs important for deployment. Modern deep forecasting models incorporate exogenous information through cross-attention~\citep{wangTimeXerEmpoweringTransformers2024}, graph encoding~\citep{GCGNetGraphConsistentGenerative2025}, or MLP feature stacking~\citep{dasLongtermForecastingTiDE2023}. While effective, they typically rely on flexible feature fusion rather than structured input-driven temporal evolution. This can let models exploit input-history correlations that do not reflect the underlying input-response behavior. Attention and graph modules add computational overhead, while lightweight MLPs may lack temporal structure.

Koopman-inspired modeling provides a useful architectural bias. For autonomous nonlinear systems, Koopman theory represents nonlinear evolution as linear dynamics over \emph{observables}, i.e., scalar-valued functions of the state. Learning nonlinear lifting with linear latent transport may be easier to learn than a nonlinear latent transition directly. Recent applications such as~\citet{zhangSKOLRStructuredKoopman2025a} suggest that this provides a suitable bias for the forecasting setting.

Motivated by this, we propose \modelname{} (\textbf{S}tructured \textbf{Li}near \textbf{D}ynamics with \textbf{E}xogenous inputs), a lightweight Koopman-inspired architecture for multi-step forecasting with exogenous inputs.
Rather than implementing an exact finite-dimensional Koopman model, \modelname{} uses learned latent linear evolution as a practical forecasting bias. It reconstructs a forecast-start latent state from past targets and inputs, then rolls it forward with future-known exogenous variables through two constraints: a lag-invariant recurrent encoder and a shared linear rollout with additive exogenous forcing.

\modelname{} achieves strong forecasting performance while remaining computationally efficient. Across electricity price and retail demand benchmarks, it attains the best average rank with cost comparable to lightweight MLP architectures. Its advantage is clearest under shifted exogenous inputs, including out-of-range (OOR) values and unseen test-time input processes.
\Cref{fig:ood_sample} shows shifted-input examples from EPF-DE and Favorita. In EPF-DE, the future wind forecast falls outside the training range, and \modelname{} tracks the ground-truth price while the baseline deviates. In Favorita, the number of promoted items is OOR, and \modelname{} better tracks sales. \Cref{fig:ood_barplot} shows that \modelname{} reduces Mean Squared Error (MSE) by 20\% on OOR EPF-DE samples and Mean Absolute Error (MAE) by 30\% in Bouc-Wen transfer, where the same nonlinear system is driven by an unseen forcing process at test time. \modelname{} achieves an 18\% lower Root Mean Squared Logarithmic Error (RMSLE) than ablations without the structured encoder or linear latent rollout on OOR Favorita samples.

Our contributions are as follows: We introduce \modelname{}, a lightweight architecture for multi-step forecasting with exogenous inputs that replaces unconstrained feature fusion with a Koopman-inspired latent recurrence. Its key constraints are a lag-invariant recurrent encoder for forecast-start state reconstruction and a shared linear rollout with additive exogenous forcing. Empirically, \modelname{} achieves state-of-the-art performance across electricity price, retail demand, and nonlinear dynamical-system benchmarks, with the largest gains under shifted exogenous inputs. The code is publicly available.\footnote{\url{https://github.com/KIT-IAI-DRACOS/slide}.}

\section{Related work}

Deep learning has become a dominant paradigm for time series forecasting, and due to its practical importance, incorporating exogenous variables is an active area of research. Approaches to forecasting with exogenous variables include attention-based architectures such as TFT and TimeXer~\citep{limTemporalFusionTransformers2021,wangTimeXerEmpoweringTransformers2024}, lightweight feedforward models such as TiDE, TSMixer, and N-BEATSx~\citep{dasLongtermForecastingTiDE2023,chenTSMixerAllMLPArchitecture2023,olivaresNeuralBasisExpansion2023,oreshkinNBEATSNeuralNetwork2021}, and graph-based models such as GCGNet~\citep{GCGNetGraphConsistentGenerative2025}. In contrast to attention-, graph-, and MLP-based approaches that primarily rely on flexible feature fusion, \modelname{} imposes a constrained input-driven latent dynamics model, which we show improves accuracy and efficiency while promoting generalization under shifted input processes.

A related line of work studies structured latent models with inputs from the perspective of nonlinear system identification and control, including Koopman-Subnet~\citep{iacobDeepIdentificationNonlinear2021}, CT-Subnet~\citep{beintemaContinuoustimeIdentificationDynamic2022}, and MamKO~\citep{liMamKOMambabasedKoopman2024}, the latter combining Koopman-style latent operators with Mamba-based sequence modeling~\citep{guMambaLinearTimeSequence2024}. These models are related in spirit to our work in that they impose structure on latent dynamics, but they are primarily motivated by modeling, identification, or control rather than exogenous multi-step forecasting. \modelname{} deviates from exact Koopman theory where necessary, incorporating deliberate design choices that make the technique specifically suitable for forecasting with known future exogenous inputs. 

More broadly, \modelname{} is related to Koopman-with-input works~\citep{proctorGeneralizingKoopmanTheory2018,kordaLinearPredictorsNonlinear2018}, Koopman-inspired forecasting models that do not incorporate exogenous variables~\citep{zhangSKOLRStructuredKoopman2025a, liuKoopaLearningNonstationary2023}, and structured state space models such as S4 and Mamba~\citep{guEfficientlyModelingLong2022,guMambaLinearTimeSequence2024}.
Additional discussion of related work is provided in Appendix~\ref{app:extended_related_work}.

\section{Problem setting}

We consider a discrete-time target series of real-valued observations $\by_\tau \in \mathbb{R}^{n_y}$. At forecast origin $k$, we observe the past target trajectory $\by_{[k]}=[\by_1,\dots,\by_k]$, where $[k]$ denotes the index set $[1,\dots,k]$, and the past exogenous trajectory $\bu_{[k]}=[\bu_1,\dots,\bu_k]$ for an input $\bu_\tau \in \mathbb{R}^{n_u}$ that affects the target but is not affected by it. We also assume access to the values available at time $k$ for the next $H$ forecast steps, denoted by $\bu_{k,[H]}=[\bu_{k,1},\dots,\bu_{k,H}]$. The double index indicates forecast origin $k$ and lead time $h\in[H]$. These future exogenous values may themselves be forecasts and need not coincide with the values that are later observed. For example, the value available at origin $k$ for lead time two, $\bu_{k,2}$, may differ from both the realized value $\bu_{k+2}$ and from a later updated forecast $\bu_{k+1,1}$.

The forecasting task is to generate a point estimate from the conditional distribution of the target random variable $\mathbf{Y}_{k,[H]} = [\mathbf{Y}_{k,1}, \dots,\mathbf{Y}_{k,H}] = [\mathbf{Y}_{k+1}, \dots,\mathbf{Y}_{k+H}]$:
\begin{equation}
    \mathbb{P}\left(\mathbf{Y}_{k,[H]} \mid \mathbf{\theta}, \by_{[k]},\bu_{[k]}, {\bu}_{k,[H]}\right),
\end{equation}
where $\mathbf{\theta}$ are the model parameters. Our desired point forecast of the target is denoted by $\hat{\by}_{k,[H]}$. 

Given $D$ training samples of the form $(\by_{[k+H]}, \bu_{[k]}, \bu_{k,[H]})$, our goal is to learn the model parameters by minimizing a task-dependent risk $\mathcal{R}(\by_{k,[H]},\hat{\by}_{k,[H]})$. The exact risks are given in Appendix~\ref{app:datasets}.

\section{Background: Koopman theory}

We model the observed time-series as a nonlinear system with state variable $\bx \in \mathbb{R}^{n_x}$ at time index $\tau$, and a nonlinear transition map $f:\mathbb{R}^{n_x}\times \mathbb{R}^{n_u} \to \mathbb{R}^{n_x}$:
\begin{subequations}
\begin{align}
  \bx_{\tau+1} &= f(\bx_{\tau}, \bu_{\tau+1})\,,\\
  \by_{\tau} &= o(\bx_{\tau}) \,.
\end{align}
\end{subequations}
We assume that the model state is (in general) not fully observable. Instead, we observe an output $\by \in \mathbb{R}^{n_y}$ obtained from a nonlinear output map $o: \mathbb{R}^{n_x}\to \mathbb{R}^{n_y}$.

The Koopman operator framework~\citep{koopmanHamiltonianSystemsTransformation1931, bruntonModernKoopmanTheory2022} provides a means of representing nonlinear dynamical systems as evolving linearly in an infinite-dimensional lifted state. Let $\phi: \mathbb{R}^{n_x}  \to \mathbb{R}$ be a scalar measurement function that is an element of an infinite-dimensional Hilbert space $\mathcal{H}$. For the corresponding zero-input autonomous system, define $f_0(\bx):=f(\bx,0)$. The linear infinite-dimensional Koopman operator $\mathcal{K}: \mathcal{H} \to  \mathcal{H}$ acts on the  measurement functions and satisfies
\begin{equation} \label{eqn:koopman}
    \mathcal{K}\phi = \phi \circ f_0, \quad\forall\phi \in \mathcal{H}\,,
\end{equation}
or equivalently $\mathcal{K}\phi(\bx_\tau)=\phi(f_0(\bx_\tau))$.
In practice, we approximate $\mathcal{H}$ by a finite dimensional subspace $\mathcal{H}_{n_z} \subset \mathcal{H}$ with $\mathcal{H}_{n_z} = \operatorname{span}\left\{\phi_j \right\}_{j=1}^{n_z}$ and associated approximate operator $\mathcal{K}_{n_z}$. 

When there are inputs $\bu$, it becomes more complicated to preserve the linearity inherent to the Koopman framework. Following~\citet{IACOBKoopmanInputsNoNoise2024}, we introduce the concept of a lifted state $\bz \in \mathbb{R}^{n_z}$ where
$\bz_{\tau}  = \Phi(\bx_{\tau}) = \left[\phi_1(\bx_{\tau}), \dots,\phi_{n_z}(\bx_{\tau})\right]\,.$
We assume that 
$\Phi \circ f_0(\cdot) \in \operatorname{span}\{\Phi\}$, implying that there exists a matrix $A \in \mathbb{R}^{n_z\times n_z}$ such that 
\begin{equation}
    \Phi \circ f_0 = A\Phi.
\end{equation}
We also assume that our output map $o$ can be exactly represented by a linear combination of the observables $\Phi$, so:
\begin{equation}
    o(\cdot) = C\Phi(\cdot)\,, \quad \text{with}\,\, C \in \mathbb{R}^{n_y\times n_z}\,.
\end{equation}
While this can be achieved by including the output functions among the observables, requiring their span to also be Koopman-invariant is generally restrictive in finite dimensions.
Under these assumptions, \citet{iacobLearningKoopmanModels2025} show that there exists an exact finite-dimensional Koopman form 
\begin{equation}
    \bz_{\tau+1} = A\bz_\tau + B(\bz_\tau, \bu_{\tau+1})\bu_{\tau+1}
\end{equation}
where $A\in\mathbb{R}^{n_z\times n_z}$ and $B(\bz_\tau,\bu_{\tau+1})$ is, in general, a nonlinear state- and input-dependent map. \modelname{} uses the input-dependent Koopman form as architectural motivation and adopts a practical forecasting-oriented parameterization of the input effect.

\begin{figure}[t]
  \centering
  \includegraphics[ width=\linewidth]{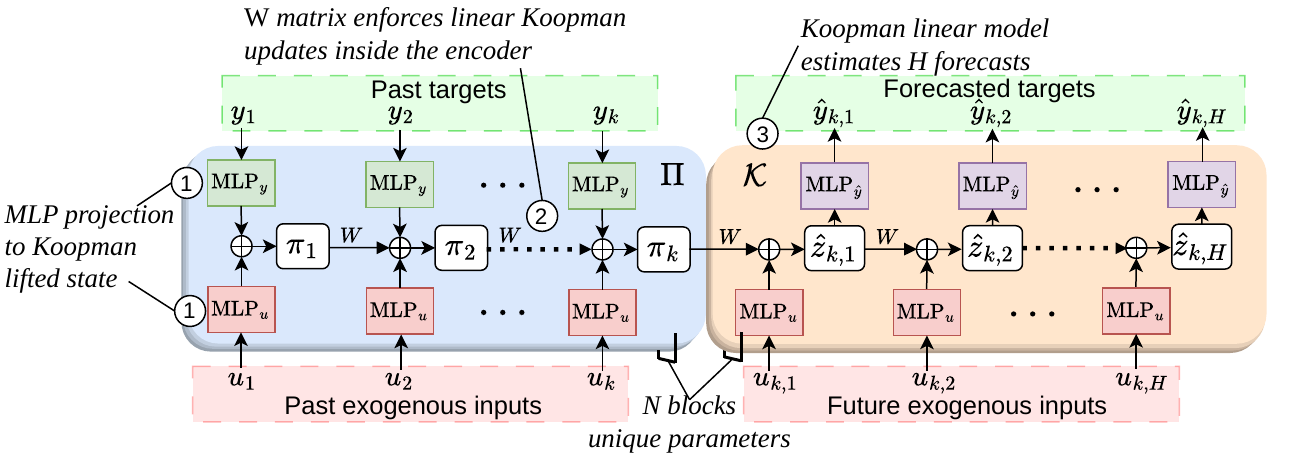}
  \caption{\modelname{} Architecture: 
  (1) MLPs project past targets and exogenous inputs to a series of hidden states $\pi_1$, \dots, $\pi_k$ inside the encoder $\Pi$. 
  (2) The encoder is parametrized by a matrix $W$, which is applied recurrently to impose linear latent dynamics. The encoder estimates the present lifted state $\pi_k = \hat{\bz}_k$.
  (3) The prediction module $\mathcal{K}$ uses $\hat\bz_k$ and future exogenous inputs to forecast the next $H$ targets. It shares the same matrix $W$ and exogenous input projection map, with an additional MLP to project back to the target dimension. $N$ independently parametrized \modelname{} blocks are jointly trained to learn complementary dynamics. The final output is taken as the sum of block forecasts.} 
  \label{fig:architecture}
\end{figure}

\section{Methodology} \label{sec:methodology}
In \modelname{}, a structured encoder maps past targets and exogenous inputs to a forecast-start latent state, which a shared linear operator $W$ rolls forward using additive forcing from future-known inputs (see \cref{fig:architecture}). A nonlinear output map maps the latent trajectory back to the target space, and multiple block forecasts are summed. We now proceed with the design principles, first motivating them based on Koopman theory, and then introducing the important adjustments for forecasting success.

\subsection{Koopman-inspired forecasting formulation}
Following~\citet{iacobLearningKoopmanModels2025} we can write our model as 
\begin{subequations}
\begin{align} \label{eq:koopman_update_definition}
    \bz_{\tau+1} &= A\bz_{\tau} + \mathcal{B}\bu_{\tau+1}\,, \\
    \by_{\tau} &= C\bz_{\tau} \,.
\end{align}
\end{subequations}
Here the matrix-valued input function $\mathcal{B}$ may be chosen simpler than in the exact formulation $B(\bz_\tau,\bu_{\tau+1})$ to trade off accuracy with model simplicity e.g. by choosing a linear, bilinear or input affine model structure~\citep{iacobLearningKoopmanModels2025}. Allowing an unrestricted lifted state dependence in $\mathcal{B}$ could introduce nonlinear dynamics in the lifted state via the input term. In our case we restrict the input function $\mathcal{B}$ to be independent of the lifted state (Appendix~\ref{app:departure} provides further context on the implications of our choice). We use this model as a motivation and treat it as a structural prior for forecasting rather than aiming at finding an exact Koopman realization.

One challenge of implementing this model is determining the initial condition of the lifted state $\bz_k$. \citet{iacobLearningKoopmanModels2025} introduce a reconstructability concept and discuss its relationship with local observability.
The {\em observability map} $\Lambda_{\bz,m}$ can be used to compute the lifted state at the start of a window from $m$ consecutive output values and the corresponding inputs. Relying on this, a {\em reconstructability map} $\Pi_{\bz,k}$ can be derived to determine the initial condition $\bz_k$ from $k$ {\em past} (lagged) outputs $\by_{[k]}$ and the $k-1$ inputs that follow the first observation, given sufficiently large $k$. We provide the derivation in the context of our method in Appendix \ref{app:derivation}.

In order to estimate this model in practice, we parameterize $A_\theta,\mathcal{B}_\theta,C_\theta$. We collect these parameters in $\theta \in \Theta \subseteq \mathbb{R}^{n_\theta}$. Since an analytical solution for $\Pi_{\bz,k}$ is impractical, we introduce a learned approximation ${\Pi}_{\bz,k}^\psi$ with parameters $\psi \in \Psi \subseteq \mathbb{R}^{n_\psi}$, which additionally receives $\bu_1$ for alignment (Appendix~\ref{app:derivation}). All parameters are learned jointly with the goal of minimizing a general risk function $\mathcal{R}(\by_{k,[H]},\hat{\by}_{k,[H]})$.

In practice, the future values $\bu_{k,h}$ available at forecast origin $k$ allow us to estimate future lifted states $\hat{\bz}_{k,h}$ using the same indexing. Our estimation approach for multi-step forecasting is then:
\begin{subequations} \label{eq:est_appr}
\begin{align} 
\hat{\bz}_{k} &= \Pi_{\bz,k}^\psi(\by_{[k]},\bu_{[k]}) \,, \\ \label{eq:matrix_koop} 
\hat{\bz}_{k,h} &= A_\theta \hat{\bz}_{k,h-1} + \mathcal{B}_\theta {\bu}_{k,h}\,, \quad \quad\text{with initialization }\hat{\bz}_{k,0} := \hat{\bz}_{k}\,,\\ 
\hat{\by}_{k,h} &= C_\theta \hat{\bz}_{k,h},\, h\in \{1,\dots,H\}\,. \end{align}
\end{subequations}

\subsection{Defining a structure on the reconstructability map}\label{sec:encoder_structure}

Beyond establishing its theoretical existence, prior work gives little guidance on the practical form of ${\Pi}_{\bz,k}$ and typically treats it as a generic black-box encoder, implemented as a dense MLP learned directly from data~\citep{iacobLearningKoopmanModels2025}. For forecasting, this approach has a number of practical challenges: a dense MLP ignores the sequential structure of the past trajectory, its parameterization grows with the available context length, and its computational cost can dominate the relatively lightweight latent rollout. We therefore impose additional structure on the learned encoder ${\Pi}_{\bz,k}^\psi$, introducing a forecasting-oriented inductive bias for reconstructing the forecast-start latent state:

{\bf Lag-invariant parameterization.} We constrain ${\Pi}_{\bz,k}^\psi$ to have the same parameterization $\psi \in \Psi \subseteq \mathbb{R}^{n_\psi}$ for any positive $k$. The parameter search space $\Psi$ is fixed as a design choice and does not grow with the number of lagged values $k$. This prevents the model search space from exploding and reduces model memory cost. Furthermore, it allows for a single $\psi$ to be learned from training samples with varying amounts of past data available. 

{\bf Encoder Koopman update.} 
The exact reconstructability map is recursively related to the latent dynamics: reconstructing $\bz_k$ from history can be viewed, informally, as reconstructing an earlier latent state and propagating it forward. Motivated by this structure, we impose a recurrent latent update inside the encoder. In particular, we define
\begin{subequations}
\begin{align}
    &\Pi_{\bz,k}^\psi(\by_{[k]},\bu_{[k]}) = \hat{\bz}_k \\
    &\,= A_\psi \hat{\bz}_{k-1} + \mathcal{B}_\psi \bu_k + \mathcal{D}_\psi\by_k \\
    &\,=  A_\psi\Pi_{\bz,k-1}^\psi(\by_{[k-1]},\bu_{[k-1]}) + \mathcal{B}_\psi \bu_k + \mathcal{D}_\psi\by_k
\end{align}
\end{subequations}
for some linear matrix $A_\psi$ and MLPs $ \mathcal{B}_\psi$ and $\mathcal{D}_\psi$. This recurrence requires the encoder to learn not just a final estimate $\hat{\bz}_k$, but a trajectory of estimates that is updated with the same linear structure as the underlying Koopman system. In this context, $\by$ is modeled as an additive input term alongside $\bu$, helping to provide the information required to guide the trajectory of $\hat{\bz}$ towards a final estimate $\hat{\bz}_k$. 

\textbf{Parameter sharing.} In order to limit the parameterization of our model and align the dynamics inside the encoder with those used in the forecast horizon, we impose complete parameter sharing by setting $A_\psi=A_\theta$ and $\mathcal B_\psi=\mathcal B_\theta$. This imposes a strong heuristic for the encoder to follow the dynamics of the Koopman model. However, the inclusion of the $\mathcal{D}_\psi\by_k$ term allows for some differences, so that the known lagged values $\by_{[k]}$ can influence the reconstruction of $\hat{\bz}_k$.

\subsection{Model definition} \label{ssec:model}
\textbf{Pre-processing: Patching.}
For long input histories or forecast horizons, we may optionally apply non-overlapping patching to the target and input sequences, reducing the effective sequence length while increasing the per-step target and input dimensions. The exact transformation is given in Appendix~\ref{app:meth_details}. After patching, we reuse the notation $\by$ and $\bu$ for the transformed target and input sequences.

\textbf{Implementation of the Koopman operator.} We implement the Koopman operator approximation as a matrix $A_{\theta} = A_{\psi} := W \in \mathbb{R}^{n_z \times n_z}$, with parameter sharing as imposed in~\cref{sec:encoder_structure}. For the input maps $\mathcal{B}_{\theta}$ = $\mathcal{B}_{\psi}$ and $\mathcal{D}_{\psi}$, we approximate the terms
\begin{subequations}
    \begin{align}
        \mathcal{B}_{\theta}\bu_{\tau}&\approx \text{MLP}_{\bu}(\bu_{\tau}) = \mathbf{p}_{\tau} \\
        \mathcal{D}_{\psi}\by_{\tau}&\approx \text{MLP}_{\by}(\by_{\tau}) = \mathbf{d}_{\tau}
    \end{align}
\end{subequations} The MLPs have much more representational ability than a fixed matrix but are less general than $B(\bz_\tau,\bu_\tau)\bu_\tau$, since the MLP output does not vary with $\hat{\bz}_\tau$. In time series forecasting tasks, the external input provided to the model may be noisy or uncertain. This choice provides a practical compromise. By alleviating restrictions on the input term, we can better match the complexity of real-world data. The resulting forecast rollout then becomes
\begin{equation}
    \hat{\bz}_{k,h} = W\hat{\bz}_{k,h-1} + \mathbf{p}_{k,h}, \quad \text{for } h\in\{1,\dots,H\}.
    \label{eq:modelevolution}
\end{equation}

\textbf{Definition of the output map.} Although the idealized formulation uses $\hat{\by}_{k,h}=C_{\theta}\hat{\bz}_{k,h}$, we implement the output map as the nonlinear function $\text{MLP}_{\hat{\by}}(\hat{\bz}_{k,h})$. This added expressivity helps compensate for approximation error in the Koopman-inspired model.

\textbf{Definition of Encoder ${\Pi}_{\bz,k}^\psi$.} We encode the initial current lifted state $\bz_k$ from the $k$ lagged values of the past trajectory. Using the previously defined $\text{MLP}_{\bu}(\bu_{\tau}) = \mathbf{p}_{\tau}$ and $\text{MLP}_{\by}(\by_{\tau}) = \mathbf{d}_{\tau}$, we define the encoder ${\Pi}_{\bz,k}^\psi$ using a linear recurrent neural network with a hidden state $\bpi \in \mathbb{R}^{n_{z}}$, where
\begin{subequations}
\begin{align}
    \bpi_{0} = \mathbf{0}, \qquad
    \bpi_{\tau} &= W\bpi_{\tau-1} + \mathbf{d}_{\tau} + \mathbf{p}_{\tau},\quad \tau \in \{1,\dots,k\}\,, \\
    {\Pi}_{\bz,k}^\psi(\by_{[k]},\bu_{[k]}) &= \hat{\bz}_{k} := \bpi_{k} \,. 
\end{align}
\end{subequations}
The equality $\hat{\bz}_k = \bpi_k$ only holds at time $k$. We can write the closed-form equation for computing $\hat{\bz}$ as repeated applications of the matrix $W$ on the encoded $\by$ and $\bu$ lags:
\begin{align}
    {\Pi}_{\bz,k}^\psi&(\by_{[k]},\bu_{[k]}) = \hat{\mathbf{z}}_{k} := \mathbf{d}_{k} + \mathbf{p}_{k} + \sum_{\tau=1}^{k-1} (W)^\tau(\mathbf{d}_{k-\tau} +\mathbf{p}_{k-\tau})
\end{align}
\textbf{Block structure.}
 We impose a block structure on the entire forecasting model by decomposing the lifted state into $N$ non-interacting subspaces. Each \modelname{} block $b\in\{1,\dots,N\}$ 
 is intended to capture different dynamics of the target system. We consider our full lifted state $\bz$ to be separable into $N$ sub-states $\bz^b \in \mathbb{R}^{n_{z,N}}$ where 
 $\bz = [{\bz^1}^\top,\dots,{\bz^N}^\top]^\top$
 and $n_{z,N}= \frac{n_z}{N}$. Each subspace is initialized by a separate encoder ${\Pi}_{\bz^b,k}^\psi$. We enforce a block-diagonal structure: $A_\theta= \operatorname{diag}(A_{\theta,1},\dots,A_{\theta,N})\,,$
which allows $\bz^b$ to evolve independently through time. A separate output MLP is used to obtain $\hat{\by}^b_{k,[H]}$, with the block forecasts summed to obtain the final output: $\hat{\by}_{k,[H]} = \sum_{b=1}^N \hat{\by}^b_{k,[H]}.$

\section{Experimental results}
\label{sec:results}

In this section, we evaluate \modelname{} along three axes:
\emph{(1) Forecasting performance} on real-world and synthetic benchmarks;
\emph{(2) Generalization under shifted exogenous inputs}, including out-of-range real-world samples and the Bouc--Wen transfer setting; and
\emph{(3) Efficiency}, assessing the tradeoff between predictive performance and computational cost.

\textbf{Benchmarks.} 
We evaluate \modelname{} on real-world datasets with heterogeneous covariates, including future-known, past-only, and static variables. In \modelname{}, future-known covariates are treated as exogenous inputs $\bu$, past-only covariates are modeled jointly with the target history, and static covariates are embedded and concatenated repeatedly to the input signal $\bu$.

The \emph{Electricity Price Forecasting (EPF) benchmark}~\citep{lagoForecastingDayaheadElectricity2021} consists of five datasets from different electricity markets and is a standard benchmark for forecasting with future-known covariates~\citep{GCGNetGraphConsistentGenerative2025,wangTimeXerEmpoweringTransformers2024,olivaresNeuralBasisExpansion2023}. The \emph{Favorita sales dataset}~\citep{KaggleFavorita} targets large-scale retail demand forecasting with promotions and holidays known in advance.
We additionally consider the synthetic \emph{Bouc-Wen hysteretic system}~\citep{schoukensThreeBenchmarksAddressing2017, boucwendataset}, a nonlinear dynamical model with exogenous forcing. We use it both as a controlled forecasting task and as a test of generalization under shifted input processes. In the \emph{Multisine} setting, models are trained and evaluated on a random-phase multisine forcing trajectory. In the \emph{Transfer} setting, models are trained on a sine-sweep forcing process and evaluated on a multisine forcing process at test time. Appendix~\ref{app:datasets} provides additional dataset details.

\textbf{Metrics.} Performance is measured by Mean Absolute Error (MAE) and Mean Squared Error (MSE) for the EPF benchmark and the Bouc-Wen system.  For the Favorita sales dataset, we report the Root Mean Squared Logarithmic Error (RMSLE). 
To assess statistical significance, we perform pairwise one-sided tests on forecast-error differences between models. We use the Diebold--Mariano test for all datasets except Favorita, for which we use a Wilcoxon test on per-series error differences. Appendix~\ref{app:stats} provides additional details and bootstrapped confidence intervals. To compare computational cost, we report the mean inference time and the trainable parameter count for each model.

\textbf{Baselines.} We compare \modelname{} against two baseline groups: competitive forecasting architectures that support exogenous information, and related Koopman- or system-identification-inspired models. The forecasting baselines are TSMixer~\citep{chenTSMixerAllMLPArchitecture2023}, TiDE~\citep{dasLongtermForecastingTiDE2023}, TimeXer~\citep{wangTimeXerEmpoweringTransformers2024}, GCGNet~\citep{GCGNetGraphConsistentGenerative2025}, and DLinear~\citep{zengAreTransformersEffective2023} as a simple linear baseline. We also include SKOLR~\citep{zhangSKOLRStructuredKoopman2025a}, a related Koopman-inspired model that cannot use future-known covariates to show the performance lift from auxiliary covariates.
To position \modelname{} relative to structured latent-dynamics approaches, we compare against Koopman-Subnet~\citep{iacobDeepIdentificationNonlinear2021}, CT-Subnet~\citep{beintemaContinuoustimeIdentificationDynamic2022}, and MamKO~\citep{liMamKOMambabasedKoopman2024}, which are conceptually relevant Koopman-with-input or system-identification baselines, though not originally proposed for the exogenous forecasting setup studied here.
Appendix~\ref{app:baselines+train_method} provides additional
context.

\textbf{Training Methodology.} To ensure a fair comparison, we select the hyperparameters for all baselines using the same random-search protocol with a fixed budget and patience. For each model, we select the hyperparameter configuration that achieves the lowest validation loss. Each model is then trained five times with different random seeds and the results are averaged. Additional details on the search spaces and training settings are provided in Appendix~\ref{app:baselines+train_method}.

\begin{table*}[ht!]
    \centering
    \caption{Performance on real-world and synthetic benchmarks, including the main ablations. The reported metrics are averaged across five random seeds. We highlight the \textbf{\first{first}}, \second{\underline{second}} and \third{third} results for each metric across baseline comparisons and the \textbf{\first{first}} across ablations. The rightmost column reports the average rank across metrics for the baselines and a hypothetical ranking for ablations. \modelname{} achieves the best average rank overall, and the ablations show that replacing the structured encoder or the linear latent rollout degrades performance. 
    }
    \label{tab:results}
    \scriptsize
    {\setlength{\tabcolsep}{2.7pt}   %
    \begin{tabular}{lccccccccccccccc|c}
    \toprule
     & \multicolumn{10}{c}{\textbf{EPF Benchmark}} & \textbf{Favorita} & \multicolumn{4}{c}{\textbf{Bouc-Wen System}} & \textbf{Avg.}\\\cmidrule(lr){2-11}\cmidrule(lr){12-12}\cmidrule(lr){13-16}
       \textbf{Model}  & \multicolumn{2}{c}{\textbf{NP}} & \multicolumn{2}{c}{\textbf{PJM}} & \multicolumn{2}{c}{\textbf{EPEX BE}} & \multicolumn{2}{c}{\textbf{EPEX FR}} & \multicolumn{2}{c}{\textbf{EPEX DE}} & \textbf{Sales} & \multicolumn{2}{c}{\textbf{Multisine}} & \multicolumn{2}{c}{\textbf{Transfer}} &  \textbf{Rank} \\ 
           \cmidrule(lr){2-3} \cmidrule(lr){4-5} \cmidrule(lr){6-7} \cmidrule(lr){8-9} \cmidrule(lr){10-11} 
           \cmidrule(lr){12-12} \cmidrule(lr){13-14}\cmidrule(lr){15-16}
      & MAE & MSE & MAE & MSE& MAE & MSE& MAE & MSE& 
      MAE& 
      MSE& 
      RMSLE& MAE & MSE& MAE & MSE \\\midrule
      \modelname{} & \first{\textbf{0.210}} & \first{\textbf{0.167}} & \second{\underline{0.157}} & \third{0.064} & \second{\underline{0.231}} & 0.367 & \first{\textbf{0.169}} & \third{0.329} & \first{\textbf{0.265}} & \first{\textbf{0.182}} & \first{\textbf{0.419}} & \third{0.085} & \third{0.012} & \first{\textbf{0.296}} & \first{\textbf{0.139}} & 1.87\\ \midrule
      \multicolumn{10}{l}{\textbf{\emph{Forecasting Architectures:}}}\\
      TimeXer & \third{0.220} & 0.186 & \first{\textbf{0.154}} & \first{\textbf{0.060}} & \first{\textbf{0.230}} & 0.367 & \first{\textbf{0.169}} & 0.347 & \second{\underline{0.278}} & \third{0.213} & 0.461 & 0.160 & 0.040 & \second{\underline{0.419}} & \second{\underline{0.273}} & 3.27\\
      TSMixer & \second{\underline{0.219}} & \third{0.184} & 0.158 & \second{\underline{0.063}} & \third{0.235} & \first{\textbf{0.335}} & \third{0.189} & 0.331 & \third{0.283} & 0.218 & \second{\underline{0.422}} & 0.092 & 0.013 & 0.541 & 0.480 & 3.27\\
      GCGNet & \third{0.220} & \second{\underline{0.177}} & \second{\underline{0.157}}&0.065&0.252& 0.367 & \third{0.189} & \first{\textbf{0.320}} & 0.285 & \second{\underline{0.206}} & 0.482 & 0.220 & 0.086 & \third{0.448} & \third{0.322} & 3.87\\
      TiDE & 0.243 & 0.200 & 0.165 & 0.081 & 0.256 & \third{0.362} & 0.200 & 0.353 & 0.295 & 0.220 & \third{0.424} & 0.143 & 0.031 & 0.485 & 0.385 & 5.27\\ 
      DLinear & 0.249 & 0.211 & 0.166 & 0.074 & 0.258 & 0.371 & 0.198 & 0.363 & 0.307 & 0.248 & 0.483 & 0.793 & 0.966 & 1.354 & 2.919 & 7.93\\
      SKOLR & 0.264 & 0.237 & 0.185 & 0.097 & 0.256 & 0.405 & 0.201 & 0.372 & 0.410 & 0.432 & 0.478 & 0.754 & 0.862 & 0.782 & 0.929 & 9.00\\
      \midrule
      \multicolumn{10}{l}{\textbf{\emph{Related System Identification or Control Baselines:}}}\\
      Koop.-Subnet & 0.251 & 0.214 & 0.181 & 0.089 & 0.250 & \second{\underline{0.360}} &  0.196 &  0.345 & 0.312 & 0.257 & 0.438 & \first{\textbf{0.064}} & \first{\textbf{0.007}} & 0.721 & 0.815 & 5.80\\
      CT-Subnet & 0.237 & 0.185 & 0.175 & 0.082 & 0.260 & 0.391 & 0.208 & 0.375 & 0.318 & 0.247 & 0.428 & \second{\underline{0.070}} & \second{\underline{0.010}} & 0.767 & 0.885 & 6.67\\ 
      MamKO & 0.245 & 0.213 & 0.172 & 0.079 & 0.256 & 0.374 & 0.192 & \second{\underline{0.324}} & 0.325 & 0.276 & 0.489 & 0.464 & 0.340 & 0.944 & 1.722 & 7.40\\ 
      \midrule
      \multicolumn{10}{l}{\textbf{\emph{Ablations:}}}\\
      Base (\modelname{}) & \first{\textbf{0.210}} & \first{\textbf{0.167}} & \first{\textbf{0.157}} & \first{\textbf{0.064}} & \first{\textbf{0.231}} & 0.367 & \first{\textbf{0.169}} & \first{\textbf{0.329}} & \first{\textbf{0.265}} & \first{\textbf{0.182}} & \first{\textbf{0.419}} & 0.085 & 0.012 & \first{\textbf{0.296}} & \first{\textbf{0.139}} & 1.87\\
      Nonlinearity & 0.214 & 0.170& 0.158 & 0.067 & 0.236 & 0.371 & \first{\textbf{0.169}} & \first{\textbf{0.329}} & 0.269 & 0.186 & 0.428 & \first{\textbf{0.035}} & \first{\textbf{0.002}} & 0.741 & 0.877 & 2.73\\
      MLP encoder & 0.223 & 0.172 & 0.160 & 0.065 & 0.239 & \first{\textbf{0.352}} & 0.183 & 0.332 & 0.281 & 0.203 & 0.423 & 0.091 & 0.014 & 0.360 & 0.207 & 2.47\\
      \bottomrule
    \end{tabular}}
\end{table*}

\subsection{Key results}

\begin{figure}[t]
    \centering
    \begin{subfigure}[t]{0.4\textwidth}
        \includegraphics[width=\textwidth, trim=70 10 140 10, clip]{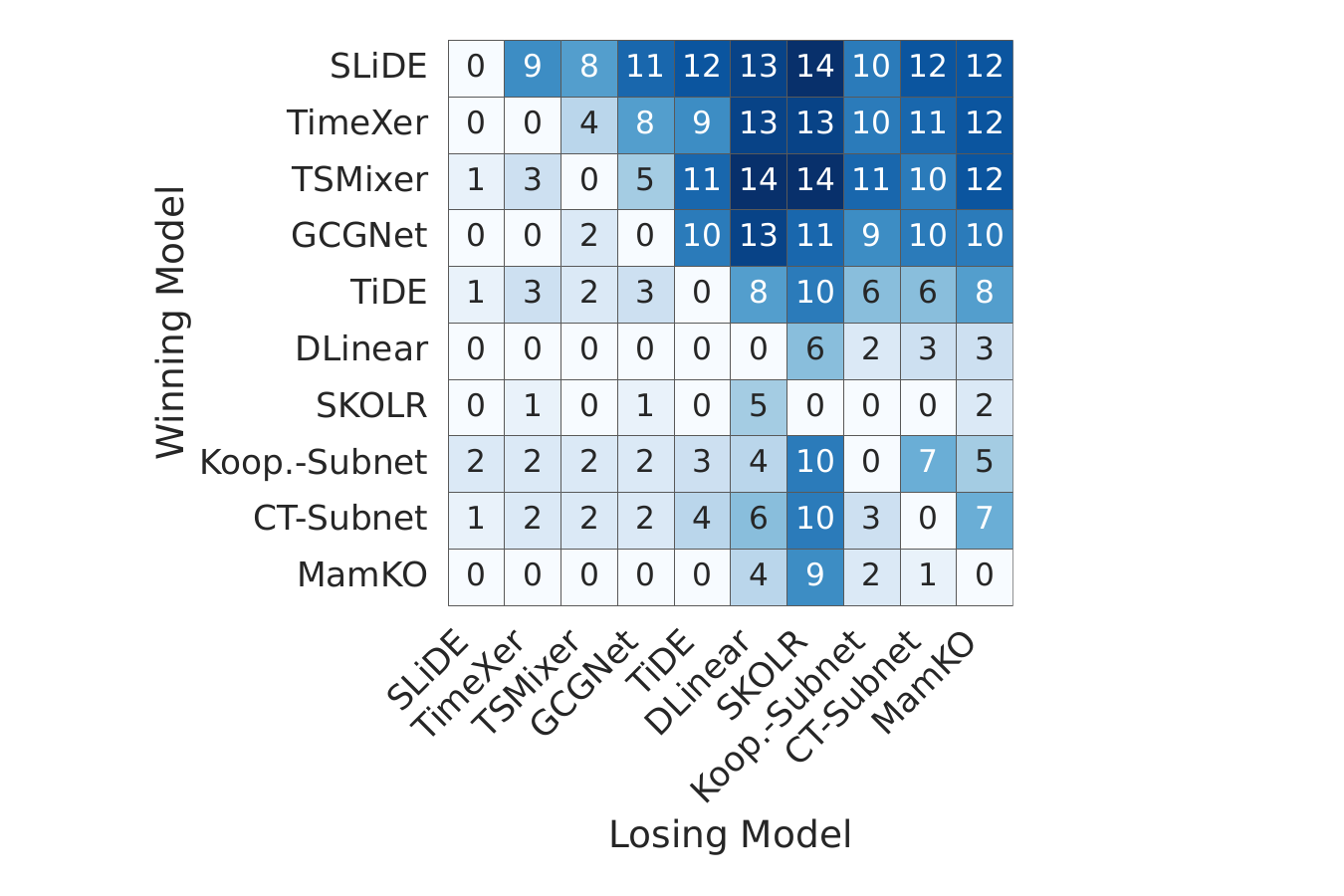}
        \caption{Pairwise significance results.}
        \label{fig:dm_results}
    \end{subfigure}
    \hspace{5em}
    \begin{subfigure}[t]{0.45\textwidth}
        \includegraphics[width=\textwidth]{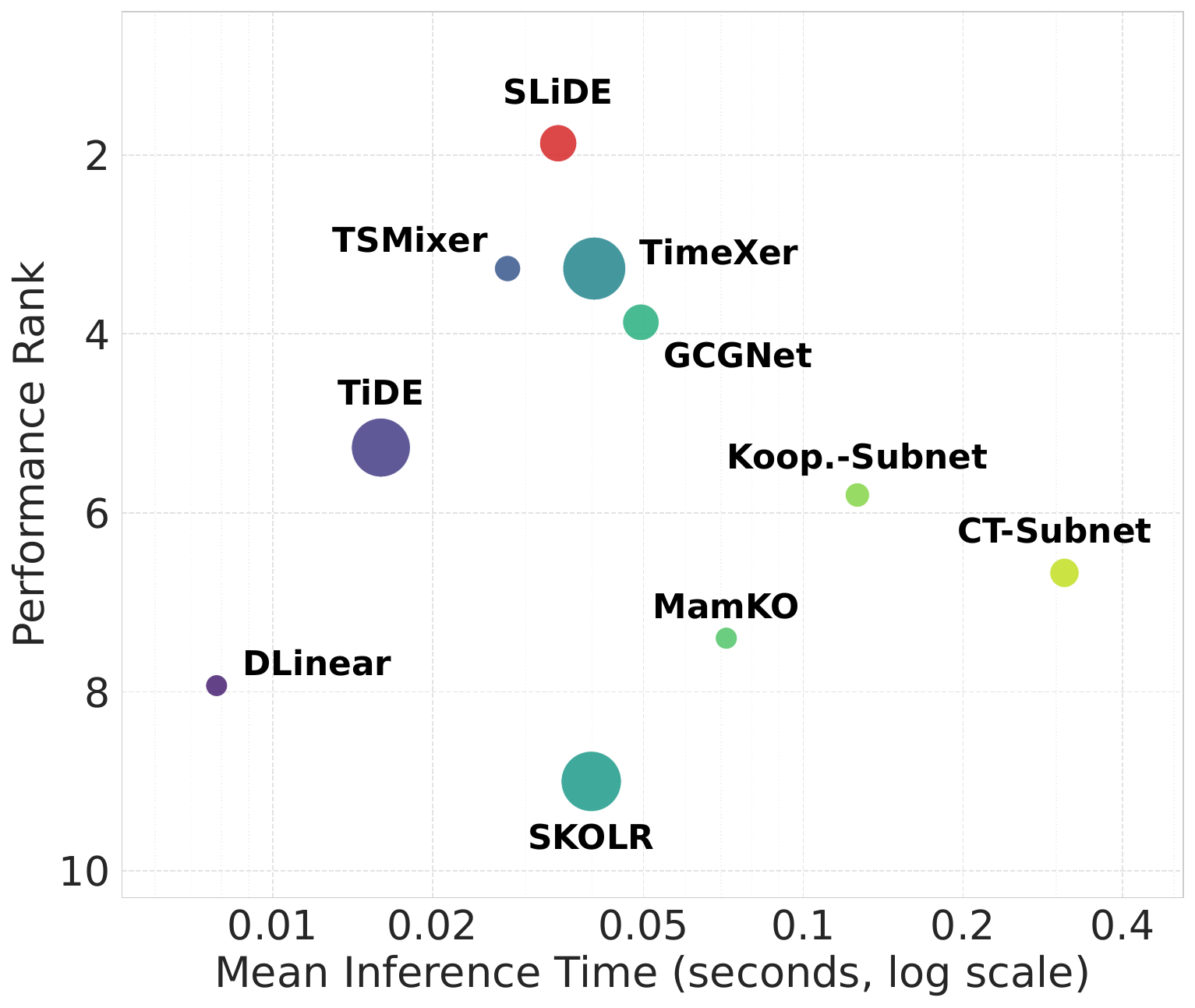}
        \caption{Efficiency-performance tradeoff.}
        \label{fig:efficiency}
    \end{subfigure}
    \caption{
    \textbf{(a)}~Pairwise statistical test results across benchmark--metric combinations. Each cell reports how often the model on the row significantly outperforms the model on the column.
    \textbf{(b)}~Mean inference time versus average performance rank; marker size is proportional to the number of trainable parameters.
    \modelname{} achieves a notably higher ranking without compromising on computational 
    cost.}
    \label{fig:dm_efficiency}
\end{figure}

\textbf{Accuracy.}
\modelname{} is a consistent top performer on the real-world datasets, ranking among the top two in eight out of eleven instances. On the synthetic Bouc-Wen benchmark, it achieves the strongest overall performance across forecasting architectures in the multisine task and the strongest performance overall in the transfer setting. While the related system identification and control baselines achieve top performance on the Bouc-Wen multisine task, they are not competitive with the forecasting architectures in general, achieving at best an average rank of 5.80. We report the benchmark results in Table~\ref{tab:results}.
Overall, \modelname{} significantly outperforms each baseline multiple times while being significantly outperformed at most twice (see~\cref{fig:dm_results}).

\textbf{Shifted Exogenous Inputs.}
\modelname{} performs particularly well under shifts in the exogenous inputs. On the real-world side, we study \emph{out-of-range} (OOR) exogenous inputs, defined as test samples for which at least one future-known exogenous value lies outside the range observed during training. We focus our analysis on EPF-DE as it contains the largest number of such samples among the EPF datasets. \Cref{fig:ood_barplot} summarizes the aggregate results: on OOR EPF-DE samples, \modelname{} reduces MSE by 20\% relative to the strongest baseline. We observe a similar pattern on Favorita, where we observe a clear separation from the baselines on a small subset of OOR samples (see Appendix~\ref{app:ood_details}). 
On the synthetic side, we use the Bouc-Wen transfer setting, where the test-time input process differs from training. In this setting, SLiDE reduces MAE by 30\% relative to the next best model.
These results support the view that \modelname{} generalizes better under shifted exogenous inputs.

\textbf{Efficiency.}
We compare mean inference time against average benchmark rank in Figure~\ref{fig:efficiency}. \modelname{} has inference cost similar to TSMixer, a model designed for computational efficiency, while being considerably faster than TimeXer and GCGNet. Marker size reflects the number of trainable parameters. Thus, \modelname{}'s superior performance does not come at a substantial computational cost.

\subsection{Ablation studies}
We perform targeted ablation experiments to assess the contribution of the main structural components of \modelname{}. In particular, we study the role of the structured encoder and the linear latent rollout.
For each ablation, we perform a random search over hyperparameters with fixed budget.

\textbf{Nonlinear State Update.}
Our proposed architecture deliberately enforces a linear temporal evolution in latent space. To examine whether this provides a useful inductive bias, we perform an ablation that introduces a nonlinear transformation into the latent dynamics by replacing $W$ with a two-layer MLP with ReLU activation. We sweep over different hidden dimensions to vary model complexity.

\textbf{MLP Encoder.}
We relax the constraints on the encoder ${\Pi}_{\bz,k}^\psi$ by replacing it with a dense MLP while keeping the shared latent rollout unchanged. We preserve the multiple-block structure and tune the hidden dimension and number of hidden layers as hyperparameters.

\textbf{Ablation results.}
\Cref{tab:results} reports the performance of both ablations across all datasets. Replacing the structured encoder with a dense MLP generally degrades performance, indicating that the lag-invariant recurrent encoder is an important component of the model. The effect of introducing a nonlinearity into the latent rollout is more nuanced. On the real-world benchmarks, it changes performance only modestly on the full test sets, while on the synthetic Bouc-Wen tasks, it improves the in-range multisine setting but substantially deteriorates performance in the transfer setting, where the same underlying system is driven by an unseen forcing process at test time. This is consistent with the view that the linear latent rollout provides a stronger bias toward generalization under shifted exogenous inputs, whereas the nonlinear variant increases modeling flexibility but is more prone to fitting forcing-specific patterns.

This pattern is clearest on shifted-input subsets. As shown in Figure~\ref{fig:ood_barplot} and the appendix tables, both ablations degrade on out-of-range (OOR) real-world samples. On EPF-DE OOR samples, the MLP encoder drops more sharply, while the nonlinear-update ablation also underperforms the base model. Favorita OOR samples show \modelname{} outperforming both ablations by 18\% in RMSLE, while in-range performance is mostly unaffected. Together, these results indicate that both the structured encoder and shared linear latent rollout contribute to \modelname{}'s stronger generalization when exogenous inputs move beyond the training regime.

We further examine the model design through additional input-map and encoder ablations in Appendix~\ref{app:arch-abl} and block-structure ablations in Appendix~\ref{app:block_ablation}. These experiments show that the proposed design remains competitive with more flexible alternatives and characterize its sensitivity to the chosen number of blocks. Finally, Appendix~\ref{app:robustness} evaluates robustness to imperfect exogenous inputs, showing that \modelname{} remains competitive under both missing and noisy future inputs.

\section{Conclusion and Limitations}

In this work, we introduced \modelname{}, a structured architecture for exogenous multi-step forecasting inspired by Koopman-style latent linear modeling. \modelname{} combines a structured lag-invariant recurrent encoder for reconstructing the forecast-start latent state with a shared linear latent rollout that propagates this state under additive exogenous forcing. Empirically, \modelname{} achieves state-of-the-art performance across real-world energy and retail benchmarks while remaining computationally efficient. Its advantage is particularly clear when future exogenous inputs shift beyond the training regime, where it outperforms the considered baselines on both real-world and synthetic tasks. Overall, our results suggest that carefully structured latent-state estimation together with a shared linear latent rollout is an effective design principle for forecasting with exogenous inputs.

\textbf{Limitations.}
Our focus is on forecasting accuracy together with computational and memory efficiency. \modelname{} inherits the strengths but also the modeling assumptions of Koopman-style latent representations. Its suitability may depend on how well a forecasting problem can be described by an approximately linear latent dynamics model. This could be more restrictive for systems with richer latent structure, for example, those not well captured by low-dimensional Koopman approximations or involving continuous spectra, which we do not study systematically here.
As exogenous inputs directly influence the latent rollout, missing, noisy, or misaligned inputs affect forecast quality. Our experiments with missing or noisy inputs (see Appendix~\ref{app:robustness}) suggest that \modelname{} is relatively robust compared to the considered baselines. Moreover, we only evaluate a limited set of shifted-input scenarios, namely out-of-range covariates with a limited number of samples per dataset and a synthetic transfer setting. 
Extending the analysis to a wider range of covariate shifts and exploring more adaptive latent operators, for example, context-dependent variants as in \citet{liMamKOMambabasedKoopman2024}, are valuable directions for future work.

\begin{ack}
This research was partially funded by the Natural Sciences and Engineering Research Council of Canada (NSERC), [reference number 260250]. Cette recherche a été partiellement financée par le Conseil de recherches en sciences naturelles et en génie du Canada (CRSNG), [numéro de référence 260250]. Ce projet de recherche no 324302 est rendu possible grâce au financement du Fonds de recherche du Québec.
The authors gratefully acknowledge funding from the Helmholtz Association under grant no. VH-NG-1727 and the Helmholtz Association’s Initiative and Networking Fund through Helmholtz AI and the HAICORE@KIT partition.
The authors gratefully acknowledge the computing time provided on the high-performance computer HoreKa by the National High-Performance Computing Center at KIT (NHR@KIT). This center is jointly supported by the Federal Ministry of Research, Technology and Space and the Ministry of Science, Research and the Arts of Baden-Württemberg, as part of the National High-Performance Computing (NHR) joint funding program. HoreKa is partly funded by the German Research Foundation (DFG).
\end{ack}

During the preparation of this work, the authors used AI-assisted tools for language editing and code development. After using these tools, the authors reviewed and edited the content as needed and take full responsibility for the content of the publication.

\clearpage

\bibliography{ref}
\bibliographystyle{plainnat}

\appendix

\appendix

\section{Derivation of the reconstructability map} \label{app:derivation}

In the following section, we provide a condensed derivation of the reconstructability map from \citet{iacobLearningKoopmanModels2025} in the context of our method. We write $\bz_{\tau+1} = A\bz_\tau + \mathcal{B}\bu_{\tau+1} := \mathcal{F}(\bz_\tau,\bu_{\tau+1})$ and use $\circ_j\mathcal F$ for $j$ successive applications of $\mathcal F$. As in the main text, $\by_{\tau,[m]}=[\by_{\tau+1}^\top,\dots,\by_{\tau+m}^\top]^\top$ and $\bu_{\tau,[m]}=[\bu_{\tau+1}^\top,\dots,\bu_{\tau+m}^\top]^\top$ (see the note on $\bu$ below).

\paragraph{Observability map.}
Consider a window of $m$ consecutive outputs $\by_{\tau,[m]}$, whose first sample is $\by_{\tau+1}=C\bz_{\tau+1}$. Propagating $\bz_{\tau+1}$ through the inputs that follow gives
\begin{equation} \label{eq:y_window}
\by_{\tau+1+j} = C\,(\circ_j\mathcal F)(\bz_{\tau+1},\bu_{\tau+1,[j]}), \qquad j=0,\dots,m-1,
\end{equation}
where, for $j=0$, $\bu_{\tau+1,[0]}$ is empty and $\circ_0\mathcal F$ is the identity, so that $\by_{\tau+1}=C\bz_{\tau+1}$. Stacking these $m$ outputs gives
\begin{equation} \label{app:eq}
\by_{\tau,[m]}=\Gamma_m(\bz_{\tau+1},\bu_{\tau+1,[m-1]}).
\end{equation}
Adapting~\citet{iacobLearningKoopmanModels2025} to our indexing, local observability from $m$ output samples requires
\begin{equation}
\underbrace{\begin{bmatrix}
\Gamma_m(\bz_{\tau+1},\bu_{\tau+1,[m-1]}) \\
\Upsilon(\bz_{\tau+1})
\end{bmatrix}}_{\mathcal{O}_{\bz,m}(\bz_{\tau+1},\bu_{\tau+1,[m-1]})} = \begin{bmatrix}
\by_{\tau,[m]} \\
0
\end{bmatrix},
\end{equation}
where $\Upsilon(\cdot)$ are nonlinear constraints on the lifted state that ensure the linear dynamics $A$ are equivalent to the original nonlinear system. Inverting $\mathcal{O}_{\bz,m}$ yields the observability map
\begin{equation}
\bz_{\tau+1} = \Lambda_{\bz,m}(\by_{\tau,[m]},\bu_{\tau+1,[m-1]}),
\end{equation}
with $\Lambda_{\bz,m} : \mathbb{R}^{m n_y} \times \mathbb{R}^{(m-1) n_u} \to \mathbb{R}^{n_z}$.

\paragraph{Reconstructability map.}
Since we prefer to determine the current state from \emph{past} data, we propagate the reconstructed window-start state to the end of the window, which takes $m-1$ forward steps:
\begin{align}
\bz_{\tau+m} &= (\circ_{m-1}\mathcal F)(\bz_{\tau+1},\bu_{\tau+1,[m-1]}) \nonumber\\
&= (\circ_{m-1}\mathcal F)\big(\Lambda_{\bz,m}(\by_{\tau,[m]},\bu_{\tau+1,[m-1]}),\,\bu_{\tau+1,[m-1]}\big) \nonumber\\
&=: \Pi_{\bz,m}(\by_{\tau,[m]},\bu_{\tau+1,[m-1]}),
\end{align}
where $\Pi_{\bz,m} : \mathbb{R}^{m n_y} \times \mathbb{R}^{(m-1) n_u} \to \mathbb{R}^{n_z}$ is the reconstructability map. Reconstruction requires a sufficiently large window $m$.

\paragraph{Full-history encoder.}
To reconstruct $\bz_k$ at present time $k$ from the full past trajectory $\by_{[k]}=\by_{0,[k]}$, we take $\tau=0$ and $m=k$:
\begin{equation}
\bz_k = \Pi_{\bz,k}(\by_{[k]},\bu_{1,[k-1]}).
\end{equation}
This map does not depend on $\bu_1$, the input aligned with the first observation. Our parameterized encoder nonetheless takes the full aligned sequence $\bu_{[k]}$ so that the same input--output structure is used at every time step:
\begin{equation}
\hat{\bz}_k = \Pi_{\bz,k}^\psi(\by_{[k]},\bu_{[k]}),
\end{equation}
where $\bu_1$ is an additional input. Computing $\Pi_{\bz,k}$ analytically is infeasible because it requires inverting $\mathcal{O}_{\bz,k}$ to obtain $\Lambda_{\bz,k}$, so $\Pi_{\bz,k}^\psi$ has learnable parameters $\psi \in \Psi \subseteq \mathbb{R}^{n_\psi}$.

\section{Additional methodology details} \label{app:meth_details}

\subsection{Characterizing the departure from the exact Koopman form}\label{app:departure}

The exact Koopman form has a state- and input-dependent input map
\begin{equation}
\bz_{\tau+1}=A\bz_{\tau}+B(\bz_{\tau},\bu_{\tau+1})\bu_{\tau+1}.
\end{equation}

However, as detailed in~\cite{iacobLearningKoopmanModels2025}, it may be desirable to trade accuracy for simplicity of the model, e.g. by choosing a suitable form of the input function $\mathcal{B}$ in~\cref{eq:koopman_update_definition}. In our case, we do not only choose the input map to be state-independent, but instead of explicitly parameterizing the matrix-valued map $\mathcal{B}$, our implementation directly parameterizes the resulting whole input forcing term as $\mathrm{MLP}_{\bu}(\bu)\in\mathbb{R}^{n_z}$. The following characterizes what this input-decoupled model retains relative to the exact form.

The input-decoupled form is exact if there exists a finite-dimensional lifting $\Phi$ for which the effect of the input on the lifted dynamics is independent of the lifted state
\begin{equation}
\Phi(f(\bx,\bu))=A\Phi(\bx)+g(\bu).
\end{equation}
This structure is analogous to a Hammerstein model in the lifted coordinates, where a static nonlinear transformation of the input drives otherwise linear latent dynamics. If, instead, the exact lifted input forcing depends on both $\bz$ and $\bu$, as permitted by the general Koopman form $B(\bz,\bu)\bu$, the state-independent forcing $g(\bu)$ constitutes an approximation.

When the input-decoupled form is not exact, we can characterize the error induced specifically by removing the latent-state dependence from the input forcing. Let $B(\bz,\bu)\bu$ denote the exact lifted forcing and let $g(\bu)$ denote its state-independent approximation. We assume that their mismatch is uniformly bounded over the relevant domain $\mathcal{D}$ of lifted states and inputs:
\begin{equation}
\sup_{(\bz,\bu)\in\mathcal{D}}
\left\lVert
B(\bz,\bu)\bu-g(\bu)
\right\rVert_2
\leq
\epsilon.
\end{equation}

Furthermore, assume that the linear latent operator $A$ is Schur stable, i.e.,
\begin{equation}
\rho(A)<1.
\end{equation}
For any $\lambda$ satisfying $\rho(A)<\lambda<1$, there then exists a constant $c\geq1$ such that
\begin{equation}
\left\lVert A^h\right\rVert_2
\leq
c\lambda^h,
\qquad h\geq0.
\end{equation}

To isolate the error introduced by the state-independent forcing, consider the exact and input-decoupled lifted dynamics under the same input sequence:
\begin{subequations}
\begin{align}
\bz_{h+1} &= A\bz_h+B(\bz_h,\bu_{h+1})\bu_{h+1}, \\
\tilde{\bz}_{h+1}&=A\tilde{\bz}_h +g(\bu_{h+1}),
\end{align}
\end{subequations}
with identical initial lifted states,
\begin{equation}
\bz_0=\tilde{\bz}_0.
\end{equation}
Define the trajectory discrepancy and the corresponding one-step forcing mismatch as
\begin{subequations}
\begin{align}
\mathbf{e}_h
&:=
\bz_h-\tilde{\bz}_h,
\\
\boldsymbol{\delta}_h
&:=
B(\bz_h,\bu_{h+1})\bu_{h+1}
-
g(\bu_{h+1}).
\end{align}
\end{subequations}
By assumption,
\begin{equation}
\left\lVert\boldsymbol{\delta}_h\right\rVert_2
\leq
\epsilon
\end{equation}
for all considered state-input pairs. Subtracting the two latent dynamics gives
\begin{equation}
\mathbf{e}_{h+1}
=
A\mathbf{e}_h
+
\boldsymbol{\delta}_h,
\qquad
\mathbf{e}_0=\mathbf{0}.
\end{equation}
Unrolling this recursion yields
\begin{equation}
\mathbf{e}_h
=
\sum_{j=0}^{h-1}
A^{h-1-j}\boldsymbol{\delta}_j.
\end{equation}
Therefore,
\begin{align}
\left\lVert\mathbf{e}_h\right\rVert_2
&\leq
\sum_{j=0}^{h-1}
\left\lVert A^{h-1-j}\right\rVert_2
\left\lVert\boldsymbol{\delta}_j\right\rVert_2
\\
&\leq
c\epsilon
\sum_{j=0}^{h-1}
\lambda^{h-1-j}
\\
&=
c\epsilon
\sum_{j=0}^{h-1}
\lambda^j
\\
&=
c\frac{1-\lambda^h}{1-\lambda}\epsilon.
\end{align}
Consequently,
\begin{equation}
\limsup_{h\rightarrow\infty}
\left\lVert\mathbf{e}_h\right\rVert_2
\leq
\frac{c}{1-\lambda}\epsilon.
\end{equation}
Thus, under Schur-stable latent dynamics, a uniformly bounded error resulting from the state-independent approximation of the input forcing does not accumulate without bound with the forecast horizon.

This bound isolates the approximation introduced by replacing the exact forcing $B(\bz,\bu)\bu$ with $g(\bu)$ while keeping the same latent operator $A$. Therefore, it does not constitute a bound on the total error of the learned forecasting model. In our implementation, the latent operator is learned as $A_\theta=W$ and the forcing is directly parameterized as
\begin{equation}
g_\theta(\bu)
:=
\mathrm{MLP}_{\bu}(\bu).
\end{equation}
The exact Koopman operator $A$ is unknown in our setting. However, our spectral analysis in Appendix~\ref{app:spectrum} shows empirically that the learned $W$ is Schur stable across the considered tasks. Appendix~\ref{app:state-dep-enc-abl} further provides a specific ablation experiment, replacing \modelname{}'s input map with a state-dependent alternative that stays closer to the structure proposed by the exact model form.

\subsection{Patching}
For time series with a long history $k$ and forecast horizon $H$, we may optionally pre-process the target time series by patching together consecutive observations. We define a patch length $P$ to transform the target time series to have $k'=k/P$ past values and $H'=H/P$ future values. For $\by^\text{patch}_{\tau} \in \mathbb{R}^{ P\cdot {n_y}}$ and $\tau\in\{1,\dots, k'+H'\}$, we define:
\begin{align} \label{eq:patching}
    \by^\text{patch}_{\tau} \triangleq [\by_{P(\tau-1)+1},\dots, \by_{P\tau}]^\top\,.
\end{align}
Within each patch, the consecutive time steps and channels of $\by$ are flattened into a space $\mathbb{R}^{ P\cdot {n_y}}$. This removes the sequential relationship of consecutive observations within the patch, treating it instead as simultaneous observations of a higher-dimensional variable. The range of defined time indices is reduced by a factor of $P$. This same patching mechanism is then analogously applied for the input sequence $\bu$.

  We begin by reframing our target time series from $\by \in \mathbb{R}^{n_y}$ to $\by^\text{patch} \in \mathbb{R}^{P\cdot n_y}$ following the patching process described in~\cref{eq:patching}. The transformed target variable is equivalent to some other target $\by'$, with a different forecast horizon $H' = H/P $ and dimension $n_y' = P\cdot n_y$. Our model treats $\by^\text{patch}$ as the variable of interest for forecasting, and we can subsequently obtain $\by$ by inverting the patching process. 

\section{Additional details on ``OOR'' results}
\label{app:ood_details}

\begin{table}[ht]
    \centering
    \caption{Number of samples, here denoted as out-of-range (OOR) samples, for which an exogenous input value exceeds the training range per EPF dataset.}
    \label{tab:epf_ood}
    {\setlength{\tabcolsep}{1.5pt}
    \begin{tabular}{lc}
    \toprule
        \textbf{Dataset}  &  \textbf{\#OOR samples}  \\\midrule
       NP & 18\\
       PJM & 0\\
       EPEX BE & 16\\
       EPEX FR & 16\\
       EPEX DE & 93\\
    \bottomrule
    \end{tabular}}
\end{table}

\begin{table}[ht]
    \centering
    \caption{Detailed EPF DE OOR Results}
    {\setlength{\tabcolsep}{2pt}
    \begin{tabular}{lccccccc}
    \toprule
    \textbf{Model} & \multicolumn{2}{c}{\textbf{OOR}} & \multicolumn{2}{c}{\textbf{In-Range}} & \multicolumn{2}{c}{\textbf{overall}} \\
     & \textbf{MAE} & \textbf{MSE} & \textbf{MAE} & \textbf{MSE} & \textbf{MAE} & \textbf{MSE} \\
    \midrule
    \modelname{} & \textbf{0.305} & \textbf{0.278} & \textbf{0.254} & \textbf{0.156} & \textbf{0.265} & \textbf{0.182} \\
    TimeXer & 0.348 & 0.384 & 0.259 & 0.166 & 0.278 & 0.213 \\
    GCGNet & 0.349 & 0.356 & 0.267 & 0.166 & 0.285 & 0.206 \\
    TSMixer & 0.367 & 0.418 & 0.261 & 0.165 & 0.283 & 0.218 \\
    TiDE & 0.379 & 0.400 & 0.272 & 0.172 & 0.295 & 0.220 \\
    Koop.-Subnet & 0.392 & 0.444 & 0.291 & 0.206 & 0.312 & 0.257 \\ 
    DLinear & 0.399 & 0.471 & 0.282 & 0.188 & 0.307 & 0.248 \\
    CT-Subnet & 0.407 & 0.452 & 0.294 & 0.192 & 0.318 & 0.247 \\
    MamKO & 0.415 & 0.495 & 0.301 & 0.217 & 0.325 & 0.276 \\
    SKOLR & 0.502 & 0.698 & 0.385 & 0.360 & 0.410 & 0.432 \\
    \midrule
    Base & \textbf{0.305} & \textbf{0.278} & \textbf{0.254} & \textbf{0.156} & \textbf{0.265} & \textbf{0.182} \\
    Nonlinearity & 0.311 & 0.285 & 0.258 & 0.160 & 0.269 & 0.186 \\
    MLP Encoder & 0.348 & 0.342 & 0.263 & 0.165 & 0.281 & 0.203 \\
    \bottomrule
    \end{tabular}}
    \label{tab:epf_de_ood}
\end{table}

\begin{table}[ht]
    \centering
    \caption{Detailed Favorita OOR Results}
    {\setlength{\tabcolsep}{2pt}
    \begin{tabular}{lccccccc}
    \toprule
    \textbf{Model} & \textbf{OOR} & \textbf{In-Range} & \textbf{overall} \\
     & \textbf{RMSLE} & \textbf{RMSLE} & \textbf{RMSLE} \\
    \midrule
    \modelname{} & \textbf{0.670} & 0.406 & \textbf{0.419} \\
    CT-Subnet & 0.752 & 0.410 & 0.428 \\
    TiDE & 0.761 & 0.405 & 0.424 \\
    TSMixer & 0.805 & \textbf{0.400} & 0.422 \\
    Koop.-Subnet & 0.808 & 0.417 & 0.438 \\
    TimeXer & 1.116 & 0.415 & 0.461 \\
    MamKO & 1.179 & 0.440 & 0.489 \\
    DLinear & 1.237 & 0.427 & 0.483 \\
    GCGNet & 1.252 & 0.424 & 0.482 \\
    SKOLR & 1.288 & 0.415 & 0.478 \\ \midrule
    Base & \textbf{0.670} & 0.406 & \textbf{0.419} \\
    MLP Encoder & 0.819 & 0.401 & 0.424 \\
    Nonlinearity & 0.877 & \textbf{0.400} & 0.428 \\
    \bottomrule
    \end{tabular}}
    \label{tab:favorita_ood}
\end{table}

\begin{figure}
    \centering
    \includegraphics[width=1\linewidth]{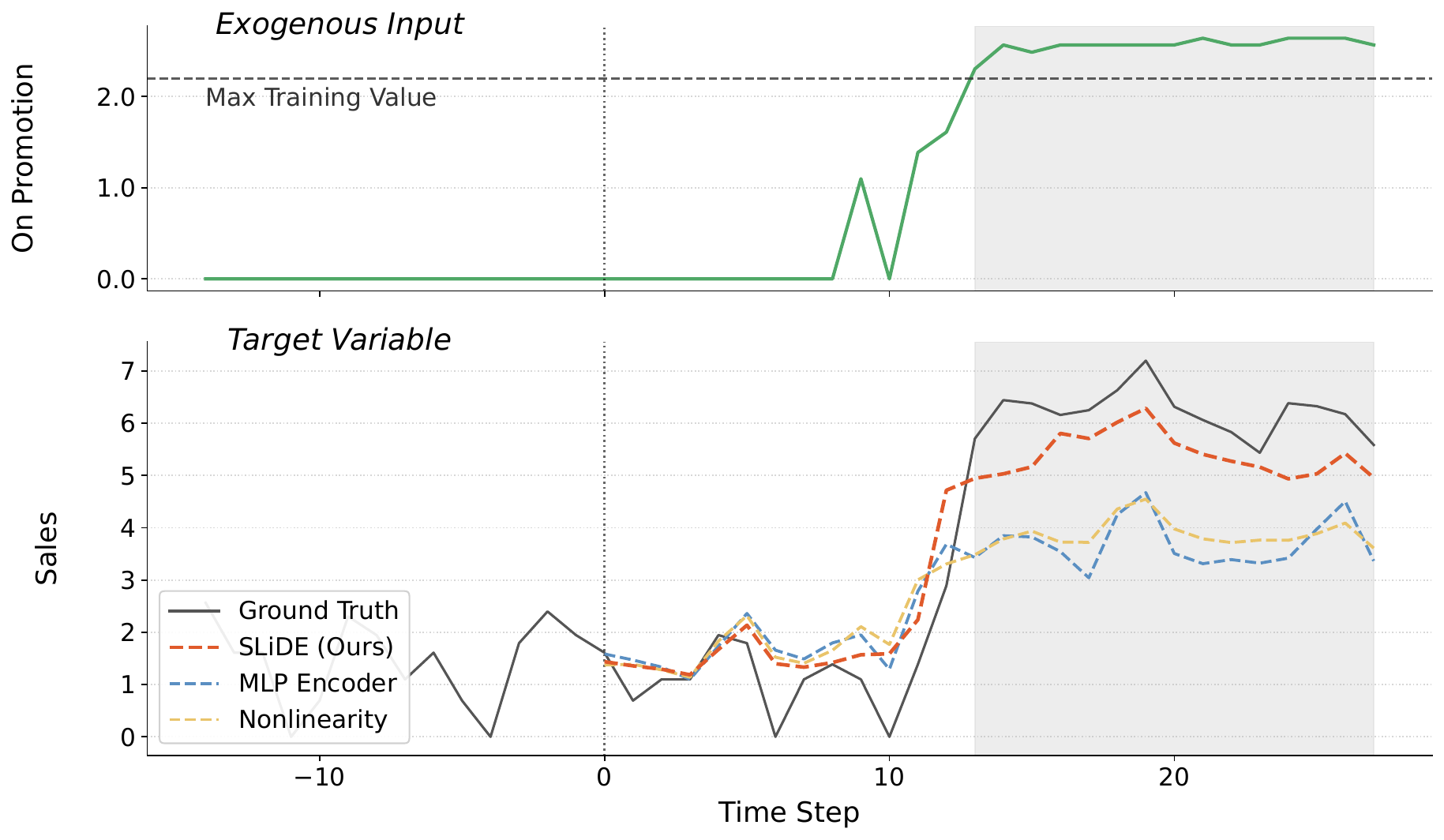}
    \caption{Favorita example with out-of-range promotion inputs. The top panel shows the future-known exogenous input, and the bottom panel the corresponding sales forecasts. Once the promotion signal exceeds the maximum training value, \modelname{} stays closer to the ground truth than either the MLP-encoder or nonlinear-rollout ablation.}
    \label{fig:favoritaablations}
\end{figure}

For the real-world datasets, \emph{out-of-range exogenous inputs} provide a concrete instance of \emph{shifted exogenous inputs}. These are test samples whose future-known exogenous values fall outside the training input regime. This setting does not imply a general distribution shift, but isolates a practical form of extrapolation in the future input process.

\paragraph{EPF datasets.}
Table~\ref{tab:epf_ood} reports the number of such OOR samples for each dataset in the EPF benchmark. The phenomenon is relatively rare in most markets, but occurs most frequently in EPF-DE, which contains 93 OOR test days, substantially more than the other EPF datasets. All EPF datasets contain 437 test days in total. We therefore focus our detailed analysis on EPF-DE, as it provides the clearest real-world setting for studying generalization beyond the training support of future-known covariates.

Table~\ref{tab:epf_de_ood} compares model performance on the OOR subset, the complementary in-range subset, and the full EPF-DE test set. \modelname{} performs best on the OOR subset, achieving the lowest MAE and MSE among all models. The margin over the strongest forecasting baselines is substantially larger on OOR samples than on the in-range subset, indicating that \modelname{}'s advantage is concentrated in cases where future exogenous inputs exceed the range seen during training. The same trend is reflected in the ablations: both the nonlinear-update and MLP-encoder variants degrade on the OOR subset, with the MLP encoder showing the larger drop. This supports the view that both the structured encoder and the shared linear latent rollout contribute to the improved behavior under shifted input conditions.

\paragraph{Favorita.}
We also perform an analogous analysis on the Favorita dataset. Here, we define a test sample as OOR if the number of promoted items for the corresponding product group, evaluated over the forecast horizon, exceeds the range observed during training for this particular group. Under this definition, 60 out of 1,579 test samples qualify as OOR. Although these samples make up only a small fraction of the test set, Table~\ref{tab:favorita_ood} shows that this is precisely the region in which \modelname{} separates most clearly from the baselines: it achieves the best RMSLE on the OOR subset, while differences on the in-range subset are much smaller. The ablation results follow the same pattern, with both the MLP encoder and nonlinear latent update degrading performance on the OOR subset. \cref{fig:favoritaablations} displays an example forecast for a test sample with out-of-range promotion inputs, where \modelname{} stays closer to the ground truth than either the MLP-encoder or nonlinear-rollout ablation. 

Although this analysis is based on a relatively small number of samples, it is consistent with the broader observation that \modelname{} is particularly effective when future exogenous inputs move beyond the support observed during training.

\paragraph{Bouc-Wen Transfer.}

\begin{figure}
    \centering
    \includegraphics[width=1\linewidth]{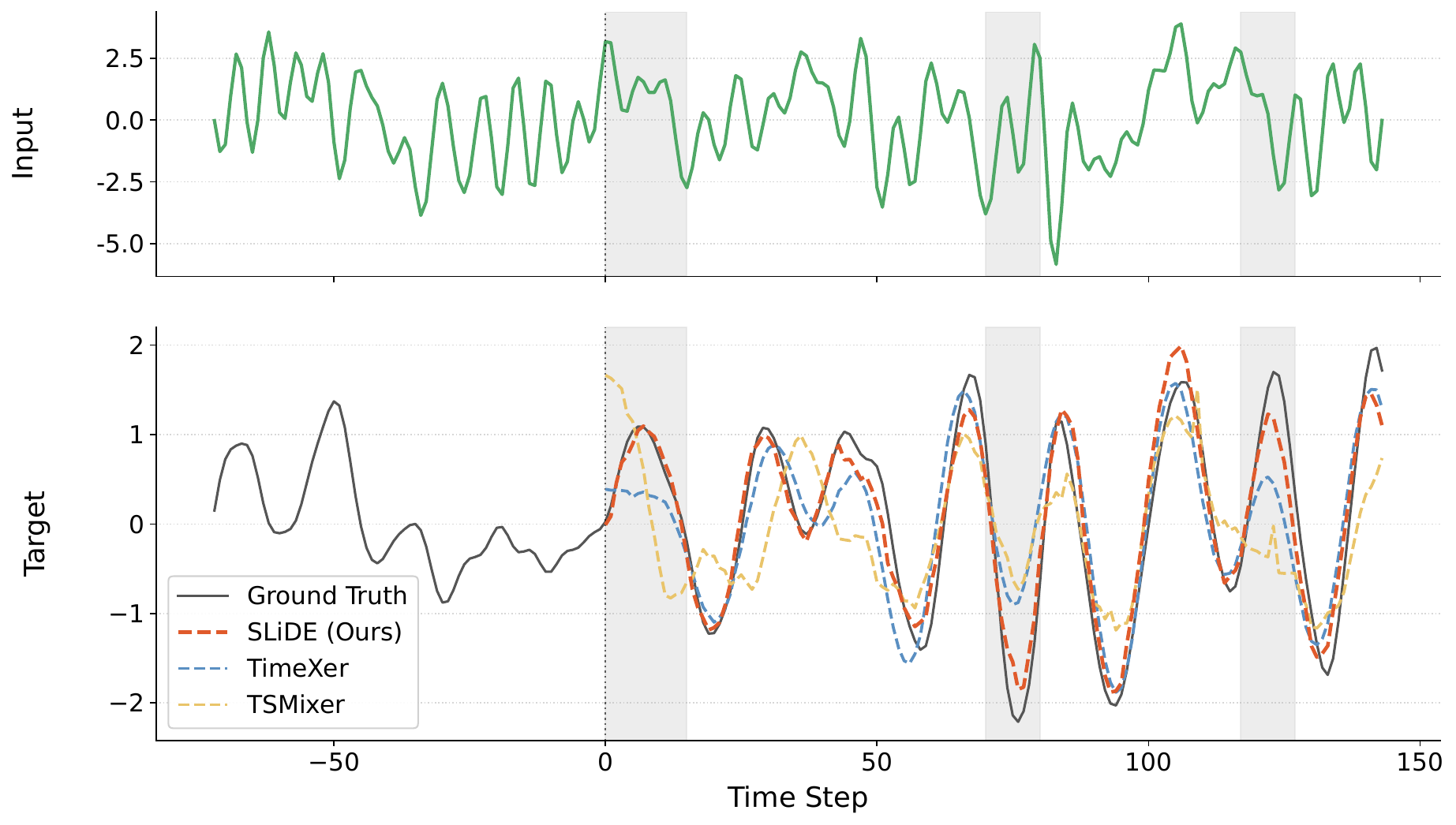}
    \caption{Illustrative Bouc-Wen transfer trajectory for \modelname{} and two baselines. The top panel shows the exogenous input process. The bottom panel compares the ground-truth target trajectory with forecasts from \modelname{}, TimeXer and TSMixer. The shaded regions highlight intervals in which \modelname{} tracks the ground-truth noticeably better than both of the baselines shown here.}
    \label{fig:bwtransferbaselines}
\end{figure}

\begin{figure}
    \centering
    \includegraphics[width=1\linewidth]{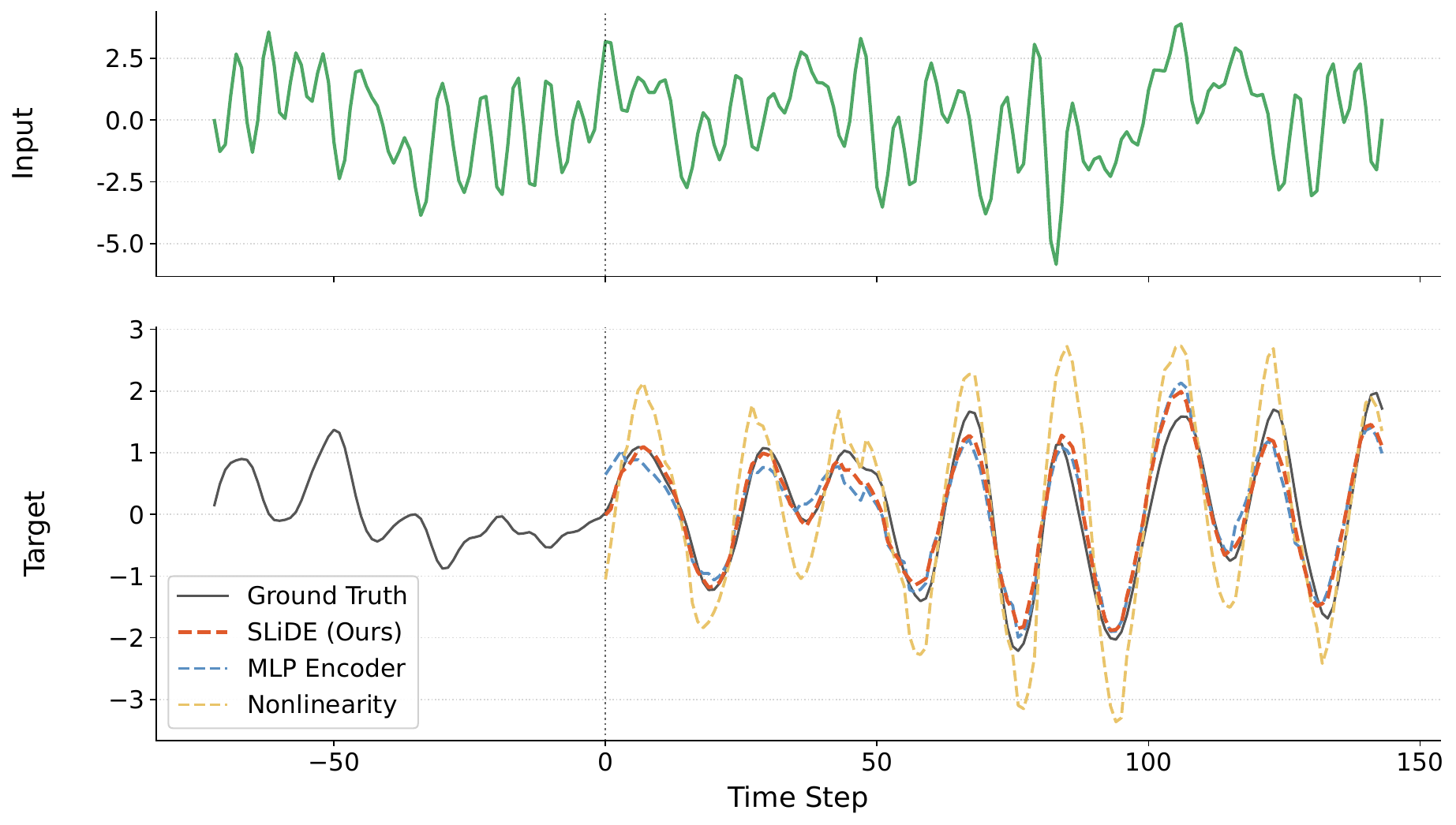}
    \caption{Illustrative Bouc-Wen transfer trajectory for \modelname{} and the two main ablations. The top panel shows the exogenous input process. The bottom panel compares the ground-truth target trajectory with forecasts from \modelname{}, the MLP-encoder ablation, and the nonlinear-rollout ablation. While the MLP-encoder ablation remains relatively close to the base model, the nonlinear latent update produces substantially larger amplitude errors.}
    \label{fig:bwtransferablations}
\end{figure}

\cref{fig:bwtransferbaselines} provides an illustrative Bouc-Wen transfer trajectory. In this example, \modelname{} remains closer to the ground truth than strong forecasting baselines across much of the rollout, especially in intervals with out-of-range input values.
\cref{fig:bwtransferablations} shows an example trajectory comparing \modelname{} to the main ablations. The nonlinear latent update produces substantially larger amplitude errors, while the MLP encoder also degrades performance relative to the base model. 

\section{Additional ablations}

\subsection{Input-map and encoder ablations}\label{app:arch-abl}
\subsubsection{State-dependent input map}\label{app:state-dep-enc-abl}

To assess the effect of removing the latent-state dependence from the
exogenous forcing, we implement a state-dependent input map following
Koopman-Subnet. An MLP takes the current latent state and input as arguments
and outputs a matrix $B(z,u)\in\mathbb{R}^{n_z\times n_u}$, such that
the additive forcing is given by $B(z,u)u$.

\begin{table*}[ht!]
    \centering
    \caption{Ablation study of the state-dependent input map and the Transformer encoder, shown together with the nonlinear state-update and MLP-encoder ablations from \cref{sec:results}. We highlight the \textbf{best} result for each metric.}
    \label{tab:add_ablations}
    \scriptsize
    {\setlength{\tabcolsep}{2.2pt}
    \begin{tabular}{lccccccccccccccc}
    \toprule
     & \multicolumn{10}{c}{\textbf{EPF Benchmark}}
     & \textbf{Favorita}
     & \multicolumn{4}{c}{\textbf{Bouc-Wen System}} \\
    \cmidrule(lr){2-11}
    \cmidrule(lr){12-12}
    \cmidrule(lr){13-16}
    \textbf{Model}
     & \multicolumn{2}{c}{\textbf{NP}}
     & \multicolumn{2}{c}{\textbf{PJM}}
     & \multicolumn{2}{c}{\textbf{EPEX BE}}
     & \multicolumn{2}{c}{\textbf{EPEX FR}}
     & \multicolumn{2}{c}{\textbf{EPEX DE}}
     & \textbf{Sales}
     & \multicolumn{2}{c}{\textbf{Multisine}}
     & \multicolumn{2}{c}{\textbf{Transfer}} \\
    \cmidrule(lr){2-3}
    \cmidrule(lr){4-5}
    \cmidrule(lr){6-7}
    \cmidrule(lr){8-9}
    \cmidrule(lr){10-11}
    \cmidrule(lr){12-12}
    \cmidrule(lr){13-14}
    \cmidrule(lr){15-16}
     & MAE & MSE
     & MAE & MSE
     & MAE & MSE
     & MAE & MSE
     & MAE & MSE
     & RMSLE
     & MAE & MSE
     & MAE & MSE \\
    \midrule
    \modelname{}
    & \textbf{0.210} & \textbf{0.167}
    & \textbf{0.157} & \textbf{0.064}
    & 0.231 & 0.367
    & 0.169 & 0.329
    & \textbf{0.265} & \textbf{0.182}
    & \textbf{0.419}
    & 0.085 & 0.012
    & \textbf{0.296} & \textbf{0.139} \\
    Nonlinearity
    & 0.214 & 0.170
    & 0.158 & 0.067
    & 0.236 & 0.371
    & 0.169 & 0.329
    & 0.269 & 0.186
    & 0.428
    & \textbf{0.035} & \textbf{0.002}
    & 0.741 & 0.877 \\
    State-dependent map
    & 0.218 & 0.174
    & 0.160 & 0.068
    & 0.234 & 0.360
    & 0.174 & 0.328
    & 0.277 & 0.207
    & 0.427
    & 0.064 & 0.007
    & 0.370 & 0.312 \\
    MLP encoder
    & 0.223 & 0.172
    & 0.160 & 0.065
    & 0.239 & \textbf{0.352}
    & 0.183 & 0.332
    & 0.281 & 0.203
    & 0.423
    & 0.091 & 0.014
    & 0.360 & 0.207 \\
    Attention-based encoder
    & 0.214 & 0.173
    & 0.159 & 0.069
    & \textbf{0.229} & 0.354
    & \textbf{0.167} & \textbf{0.318}
    & 0.267 & 0.184
    & 0.426
    & 0.075 & 0.010
    & 0.297 & 0.140 \\
    \bottomrule
    \end{tabular}
    }
\end{table*}

The state-dependent map improves on the noiseless, fully deterministic Multisine task, as does the nonlinear rollout in \cref{sec:results}, but it is worse on the real-world benchmarks and degrades sharply on the Transfer task, where the exogenous input process shifts at test time. The additional state dependence is therefore helpful only in that clean deterministic setting. The input-only forcing performs better on the real-world tasks and under shifted forcing, and it uses fewer parameters: the state-dependent MLP must output an $n_z\times n_u$ matrix rather than an $n_z$-dimensional forcing vector.

\subsubsection{Attention-based encoder}
We implement an attention-based encoder while keeping the linear rollout fixed. Tokens are embedded using the same MLPs for $y_\tau$ and $u_\tau$ and summed, then processed by a standard Transformer with RoPE self-attention.
The final hidden state is used as $\hat z_k$. Among the ablations in \cref{tab:add_ablations}, this encoder is the most competitive with \modelname{}. However, this architecture suffers from an increase in parameter count and computational cost, since self-attention replaces the linear recurrent encoder, and does not provide a clear performance uplift relative to \modelname{}.

\subsection{Block-structure ablations} \label{app:block_ablation}
\subsubsection{Selection of and sensitivity to block count}
\begin{figure}[ht]
  \centering
  \includegraphics[ width=1.0\linewidth]{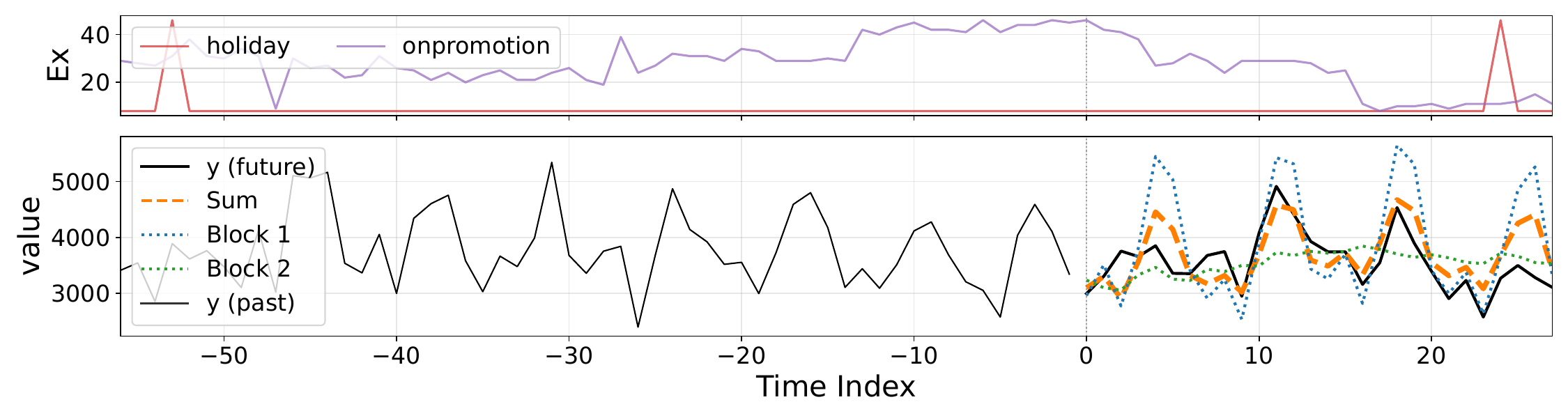}
  \caption{Forecast sample of \modelname{} on the Favorita benchmark for predicting product sales. Top: The exogenous inputs, indicating the presence of holidays and number of related products on promotion. Bottom: Sales target and \modelname{} forecast broken down into individual block forecasts. We observe that the two blocks learn different temporal patterns, which are combined to form the final forecast.}
  \label{fig:favorita_traj}
\end{figure}

\cref{fig:favorita_traj} illustrates a Favorita forecast in which the two block contributions exhibit different temporal patterns and combine to form the final prediction.
 To further investigate and inform the choice of the number of blocks $N$, we ran a dedicated ablation with $N \in {1,2,4,8}$ in two regimes: (i) fixed per-block dimension, where the total parameter count grows with $N$, and (ii) fixed total latent dimension, where more blocks make $W$ sparser and reduce parameters.

\cref{tab:block_ablation} reports the results. On average, performance improves from one to two or four blocks, with only minor differences between $N{=}2$ and $N{=}4$. Increasing the number of blocks to $N{=}8$ generally degrades performance, particularly in the fixed-total-dimension regime, where the average relative performance gap to the best-performing variant reaches $18.73\%$.
We therefore selected $N{=}2$ as the default for its favorable performance--cost trade-off and excluded $N$ from HPO to reduce computational cost.

\begin{table*}[ht]
    \centering
    \caption{Block ablation results with fixed latent dimension (\emph{Fixed per-block dimension}) and with latent dimension scaled inversely to the number of blocks (\emph{Fixed total dimension}). We highlight the \textbf{\first{first}}, \second{\underline{second}} and \third{third} results across both experiments for each metric.
    }
    \label{tab:block_ablation}
    \footnotesize
    {\setlength{\tabcolsep}{1.4pt}
    \begin{tabular}{lccccccccccccccc}
    \toprule
     & \multicolumn{10}{c}{\textbf{EPF Benchmark}} & \textbf{Favorita} & \multicolumn{4}{c}{\textbf{Bouc-Wen System}} \\

     \cmidrule(lr){2-11}\cmidrule(lr){12-12}\cmidrule(lr){13-16}

     \textbf{Blocks} & \multicolumn{2}{c}{\textbf{NP}} & \multicolumn{2}{c}{\textbf{PJM}} & \multicolumn{2}{c}{\textbf{EPEX BE}} & \multicolumn{2}{c}{\textbf{EPEX FR}} & \multicolumn{2}{c}{\textbf{EPEX DE}} & \textbf{Sales} & \multicolumn{2}{c}{\textbf{Multisine}} & \multicolumn{2}{c}{\textbf{Transfer}} \\

     \cmidrule(lr){2-3} \cmidrule(lr){4-5} \cmidrule(lr){6-7} \cmidrule(lr){8-9} \cmidrule(lr){10-11}
     \cmidrule(lr){12-12} \cmidrule(lr){13-14}\cmidrule(lr){15-16}

     & MAE & MSE & MAE & MSE & MAE & MSE & MAE & MSE & MAE & MSE & RMSLE & MAE & MSE & MAE & MSE \\
     \midrule
     \multicolumn{16}{l}{\textbf{\emph{Fixed per-block dimension:}}}\\
     1 & 0.214 & 0.173 & 0.159 & \third{0.065} & 0.235 & 0.380 & \first{\textbf{0.167}} & 0.344 & 0.269 & 0.191 & 0.423 & \first{\textbf{0.082}} & \first{\textbf{0.011}} & \third{0.306} & \third{0.148} \\
     2 & \third{0.210} & \third{0.167} & \second{\underline{0.157}} & \second{\underline{0.064}} & \second{\underline{0.231}} & 0.367 & \second{\underline{0.169}} & \third{0.329} & \third{0.265} & \second{\underline{0.182}} & \first{\textbf{0.419}} & \third{0.085} & \second{\underline{0.012}} & \second{\underline{0.296}} & \second{\underline{0.139}} \\
     4 & \second{\underline{0.209}} & \first{\textbf{0.164}} & \second{\underline{0.157}} & \second{\underline{0.064}} & \second{\underline{0.231}} & 0.366 & \second{\underline{0.169}} & \second{\underline{0.324}} & \third{0.265} & \second{\underline{0.182}} & 0.424 & 0.086 & \second{\underline{0.012}} & \third{0.306} & 0.149 \\
     8 & 0.211 & 0.169 & \first{\textbf{0.155}} & \first{\textbf{0.063}} & \first{\textbf{0.229}} & \first{\textbf{0.357}} & \third{0.172} & \first{\textbf{0.320}} & \second{\underline{0.264}} & 0.184 & 0.428 & 0.091 & 0.014 & 0.327 & 0.169 \\
     \midrule
     \multicolumn{16}{l}{\textbf{\emph{Fixed total dimension:}}}\\
     1 & \first{\textbf{0.208}} & \second{\underline{0.165}} & 0.159 & 0.066 & \third{0.232} & \second{\underline{0.361}} & \first{\textbf{0.167}} & 0.337 & 0.267 & 0.189 & \third{0.421} & \second{\underline{0.084}} & \second{\underline{0.012}} & 0.307 & 0.150 \\
     4 & 0.211 & 0.171 & \second{\underline{0.157}} & \first{\textbf{0.063}} & \third{0.232} & 0.367 & 0.173 & 0.332 & \first{\textbf{0.262}} & \first{\textbf{0.180}} & \second{\underline{0.420}} & 0.086 & \second{\underline{0.012}} & \first{\textbf{0.291}} & \first{\textbf{0.134}} \\
     8 & 0.212 & 0.173 & \third{0.158} & \first{\textbf{0.063}} & \third{0.232} & \third{0.363} & 0.176 & 0.336 & \third{0.265} & \third{0.183} & 0.424 & 0.089 & \third{0.013} & 0.447 & 0.367 \\
     \bottomrule
    \end{tabular}}
\end{table*}

\subsubsection{Explicit block diversity}

While \cref{fig:favorita_traj} suggests that different blocks learn distinct temporal patterns, \cref{fig:spectra} shows that their eigenvalue spectra largely overlap.
To investigate whether explicitly enforcing diversity improves performance, we added a diversity loss that penalizes pairwise squared cosine similarity between block operators under the Frobenius inner product, following the cosine-similarity diversity regularization of~\citet{wortsmanLearningNeuralNetwork2021}.
We evaluated weights $w{=}1$ and $w{=}10$ across $N \in \{1,2,4,8\}$.

The effect on performance is negligible, with the mean relative performance change across dataset--metric pairs ranging from $-0.36\%$ to $+0.02\%$ across the tested configurations.
Thus, explicitly encouraging block diversity does not yield systematic performance improvements in the considered experiments.

\subsection{Robustness} \label{app:robustness}
\subsubsection{Missing exogenous inputs} \label{app:missing}

\begin{figure}[ht]
    \centering
    \includegraphics[width=0.95\linewidth]{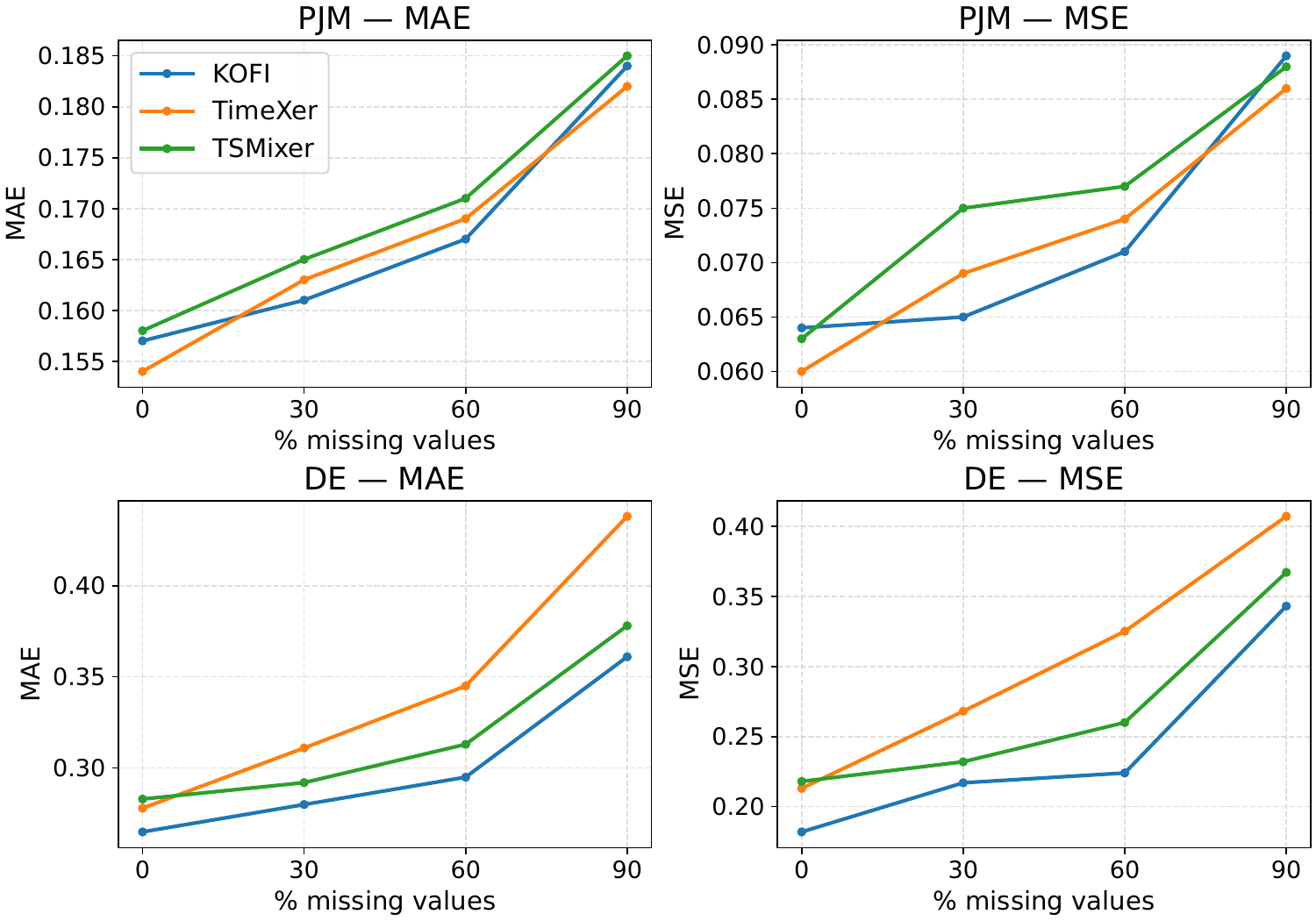}
    \caption{Evaluation metrics (MAE left, MSE right) of SLiDE, TimeXer and TSMixer against percentage of missing exogenous values for the EPF PJM (top) and DE (bottom) datasets. \modelname{} outperforms baselines on PJM with 30\% and 60\% of missing values, and is significantly more robust than TimeXer, the second-best ranking model, on EPF DE.}
    \label{fig:missing_sweep_abs}
\end{figure}

We evaluate robustness to missing exogenous inputs on EPF PJM and EPF DE, chosen to represent both a setting in which competing baselines are strong (PJM) and a setting in which \modelname{} significantly outperforms (DE). For each time step, we independently drop all exogenous values with probability $p$ according to a Bernoulli distribution. Missing values are replaced with zeros after dataset-wide standardization, and an additional binary indicator channel is appended to mark whether the exogenous values at that time step are observed or missing. We reuse the hyperparameter configuration selected in the main experiments and report average metrics over five independent runs with different random seeds. \Cref{fig:missing_sweep_abs} shows that \modelname{} outperforms TimeXer and TSMixer at 30\% and 60\% missing values on EPF PJM, despite being slightly behind at full availability. \modelname{}'s performance degradation is comparable to the lighter TSMixer, while attention-based TimeXer degrades most sharply, demonstrating strong robustness relative to the most competitive baseline available.

\subsubsection{Noisy future exogenous inputs}
We note that our real-world benchmarks already contain errors in future exogenous inputs by construction: the future-known covariates in all five EPF datasets are themselves day-ahead forecasts, not realized values. The EPF results therefore already evaluate the models using
future exogenous variables that contain realistic forecast uncertainty.

To additionally characterize robustness in a controlled way, we ran an input corruption experiment. We perturb only the future exogenous inputs in the horizon with Gaussian noise (standard deviation scaled to 50\% of each input's standard deviation), keeping past values unperturbed. Each model is re-trained with its original hyperparameters on unperturbed past and noisy future inputs. We omit Favorita from this experiment since its future-known covariates (holidays, promotions, calendar features) are deterministically known in advance, so corrupting them would not correspond to a realistic uncertainty scenario. 

\begin{table*}[ht!]
    \centering
    \caption{Performance under noisy future exogenous inputs on the EPF benchmark and Bouc-Wen tasks. 
    We highlight the \textbf{\first{first}}, \second{\underline{second}}, and \third{third}
    results for each metric.}
    \label{tab:error_robustness}
    \footnotesize
    {\setlength{\tabcolsep}{2pt}
    \begin{tabular}{lcccccccccccccc}
    \toprule
     & \multicolumn{10}{c}{\textbf{EPF Benchmark}}
     & \multicolumn{4}{c}{\textbf{Bouc-Wen System}}
      \\
    \cmidrule(lr){2-11}\cmidrule(lr){12-15}

    \textbf{Model}
     & \multicolumn{2}{c}{\textbf{NP}}
     & \multicolumn{2}{c}{\textbf{PJM}}
     & \multicolumn{2}{c}{\textbf{EPEX BE}}
     & \multicolumn{2}{c}{\textbf{EPEX FR}}
     & \multicolumn{2}{c}{\textbf{EPEX DE}}
     & \multicolumn{2}{c}{\textbf{Multisine}}
     & \multicolumn{2}{c}{\textbf{Transfer}} \\

    \cmidrule(lr){2-3}
    \cmidrule(lr){4-5}
    \cmidrule(lr){6-7}
    \cmidrule(lr){8-9}
    \cmidrule(lr){10-11}
    \cmidrule(lr){12-13}
    \cmidrule(lr){14-15}

     & MAE & MSE
     & MAE & MSE
     & MAE & MSE
     & MAE & MSE
     & MAE & MSE
     & MAE & MSE
     & MAE & MSE
      \\

    \midrule

    SLiDE
    & \first{\textbf{0.219}}
    & \first{\textbf{0.181}}
    & \third{0.163}
    & \third{0.069}
    & \first{\textbf{0.236}}
    & \third{0.369}
    & \second{\underline{0.176}}
    & 0.339
    & \first{\textbf{0.278}}
    & \first{\textbf{0.203}}
    & \first{\textbf{0.243}}
    & \first{\textbf{0.096}}
    & \first{\textbf{0.336}}
    & \first{\textbf{0.178}}
    \\

    \midrule
    \multicolumn{15}{l}{\textbf{\emph{Forecasting Architectures:}}}\\

    TimeXer
    & \second{\underline{0.222}}
    & \third{0.188}
    & \first{\textbf{0.158}}
    & \first{\textbf{0.064}}
    & \first{\textbf{0.236}}
    & 0.377
    & \first{\textbf{0.173}}
    & 0.348
    & \second{\underline{0.290}}
    & \third{0.229}
    & 0.284
    & 0.128
    & 0.503
    & 0.395
    \\

    GCGNet
    & \third{0.227}
    & \second{\underline{0.184}}
    & \second{\underline{0.162}}
    & \second{\underline{0.068}}
    & 0.252
    & 0.373
    & \third{0.190}
    & \first{\textbf{0.315}}
    & \third{0.296}
    & \second{\underline{0.221}}
    & 0.335
    & 0.178
    & \third{0.481}
    & \third{0.366}
    \\

    TSMixer
    & 0.231
    & 0.201
    & 0.165
    & 0.086
    & \third{0.242}
    & \first{\textbf{0.343}}
    & 0.200
    & \second{\underline{0.316}}
    & 0.304
    & 0.250
    & 0.286
    & 0.131
    & 0.670
    & 0.706
    \\

    TiDE
    & 0.247
    & 0.206
    & 0.164
    & 0.075
    & 0.255
    & \second{\underline{0.357}}
    & 0.201
    & 0.350
    & 0.309
    & 0.247
    & 0.281
    & 0.126
    & 0.494
    & 0.383
    \\

    \midrule
    \multicolumn{15}{l}{\textbf{\emph{Related System Identification or Control Baselines:}}}\\

    Koop.-Subnet
    & 0.265
    & 0.234
    & 0.195
    & 0.097
    & 0.259
    & 0.372
    & 0.198
    & 0.336
    & 0.334
    & 0.289
    & \second{\underline{0.250}}
    & \second{\underline{0.100}}
    & \second{\underline{0.463}}
    & \second{\underline{0.337}}
    \\

    CT-Subnet
    & 0.257
    & 0.208
    & 0.185
    & 0.089
    & 0.263
    & 0.396
    & 0.210
    & 0.386
    & 0.337
    & 0.288
    & \third{0.262}
    & \third{0.109}
    & 0.768
    & 0.885
    \\

    MamKO
    & 0.265
    & 0.242
    & 0.182
    & 0.085
    & 0.269
    & 0.395
    & 0.206
    & \third{0.334}
    & 0.375
    & 0.355
    & 0.508
    & 0.399
    & 0.744
    & 0.968
    \\

    \bottomrule
    \end{tabular}
    }
\end{table*}

SLiDE maintains its top average rank under this noisy regime. Notably, on Bouc--Wen Multisine, \modelname{} now outperforms Koopman-Subnet and CT-Subnet, which outperform \modelname{} on the task without noise addition but show particularly large relative degradation when the future forcing is corrupted. Results are reported in~\cref{tab:error_robustness}.

\section{Spectrum of the learned operator} 
\label{app:spectrum}

\begin{figure}
    \centering
    \includegraphics[width=1\linewidth]{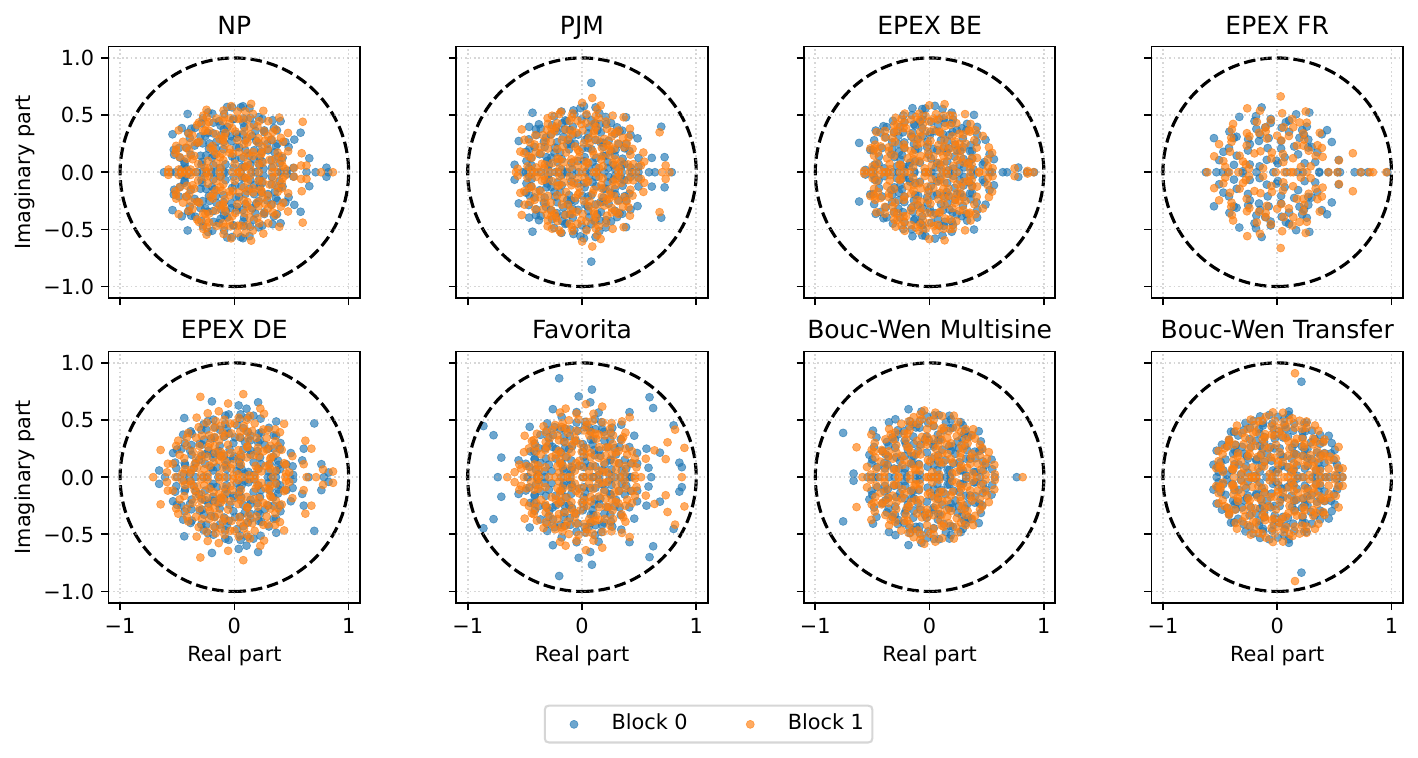}
    \caption{Eigenvalue spectra of the learned operator $W$ across blocks and datasets. Each point denotes an eigenvalue of a block-specific operator and the dashed circle indicates the unit circle.}
    \label{fig:spectra}
\end{figure}

We analyze the eigenvalue spectrum of the learned operator $W$ across blocks and datasets, as shown in~\cref{fig:spectra}. Although no explicit spectral constraint is imposed during training, all learned eigenvalues remain inside the unit circle for the considered models. Hence, the learned operators are Schur stable, empirically satisfying the stability assumption used in our error bound in App.~\ref{app:departure}. Most eigenvalues lie well within the stability boundary, indicating predominantly fast-decaying latent modes. Note that $W$ was initialized using PyTorch's default \texttt{nn.Linear} initialization, which, for sufficiently large square matrices, yields an approximately uniform eigenvalue distribution within a disk of radius $1/\sqrt{3}$.

Across blocks, the eigenvalue distributions overlap substantially, with no clear partition into distinct frequency bands or decay timescales. The block structure therefore does not appear to induce a spectral decomposition of the dynamics. Instead, the results are consistent with the blocks providing a parameter-efficient collection of parallel latent dynamics, despite the corresponding block forecasts specializing to different temporal patterns (see~\cref{fig:favorita_traj}).

The Bouc--Wen benchmark provides an informative case study. Its linearized dynamics have a reported natural frequency of $\qty{35.59}{\hertz}$~\citep{boucwendataset}. At a sampling rate of $\qty{750}{\hertz}$, this corresponds to discrete-time eigenvalue angles of approximately $\pm 2.38$ rad ($\pm 137^\circ$) for patch length $P=8$ and $\pm 1.19$ rad ($\pm 68^\circ$) for $P=4$. The dominant learned conjugate pairs roughly agree with these positions. This correspondence suggests that $W$ captures meaningful oscillatory structure of the underlying system, while the remaining, generally faster-decaying modes provide additional flexibility for representing the nonlinear hysteretic dynamics.

\section{Experimental details} \label{app:expdetails}
\subsection{Additional information on datasets} \label{app:datasets}
\begin{table}[ht]
    \centering
    \caption{Summary of the datasets and covariate types used in our experiments.}
    \label{tab:datasets}
    {\setlength{\tabcolsep}{1.4pt}
    \begin{tabular}{lccccccccc}
    \toprule
        \textbf{Dataset}  & \textbf{\# Targets} & \textbf{\# Series} & \textbf{Past} & \textbf{Future} & \textbf{Static} &  \textbf{Context} & \textbf{Horizon} & \textbf{Loss} & \textbf{Metrics}  \\\midrule
        EPF datasets & 1 & 1 & 0 & 2 & 0 & 168 &  24 & MAE & MAE, MSE\\
        Favorita Sales & 1 & 1,579 & 1 & 6 & 6 & 56 & 28 & MSE (log) & RMSLE\\
        Bouc-Wen tasks & 1 & 1 & 0 & 1 & 0 & 288 & 144 & MAE & MAE, MSE \\
        \bottomrule
    \end{tabular}}
\end{table}

\paragraph{Electricity Price Forecasting (EPF) Benchmark.} The EPF benchmark~\citep{lagoForecastingDayaheadElectricity2021} consists of five datasets of different electricity markets. For each dataset, the task is forecasting the hourly day-ahead electricity price for the next 24 hours, given two forecasted covariates for the next day. The two forecasted covariates differ for each dataset. All datasets include six years of hourly data.

\emph{Nord Pool (NP):} The NP dataset contains price data of the European power market of the Nordic countries. The two additional covariates are the day-ahead load forecast and the day-ahead wind generation forecast.

\emph{Pennsylvania–New Jersey-Maryland (PJM):} This dataset includes prices from the Commonwealth Edison (COMED) zone from the PJM market in the United States as the target time series. The covariates are two day-ahead load forecast series, a system load forecast and a zonal load forecast for the COMED zone.

\emph{European Power Exchange - Belgium (EPEX-BE):} The EPEX-BE dataset contains the day-ahead prices for the Belgian bidding zone on the market operated by EPEX SPOT. The two covariates in this dataset are the day-ahead load forecast and the day-ahead generation forecast in France. 

\emph{European Power Exchange - France (EPEX-FR):} The EPEX-FR dataset contains the day-ahead prices for the French bidding zone on the market operated by EPEX SPOT. The two covariates in this dataset are the day-ahead load forecast and the day-ahead generation forecast.

\emph{European Power Exchange - Germany (EPEX-DE):} The EPEX-DE dataset includes the day-ahead prices for the German bidding zone on the market operated by EPEX SPOT. The two covariates in this dataset are the day-ahead load forecast for the control area operated by the transmission system operator Amprion and the aggregated day-ahead wind and solar generation forecast in the zones operated by the transmission system operators Amprion, TenneT and 50Hertz.

Following~\citet{GCGNetGraphConsistentGenerative2025}, we use a 70\%/10\%/20\% train/validation/test split, a forecast horizon of 24 hours and a context length of one week (168 hours). In contrast to their evaluation setup, however, we extract inference samples with a stride of 24 hours rather than 1 hour in order to better reflect the application setting and realistic future-covariate availability. We train all models using a Mean Absolute Error (MAE) loss $\mathcal{R}
= \frac{1}{DH}\sum_{d=1}^{D}\sum_{h=1}^{H}\Big|y_{k_d+h}-\hat y_{k_d, h}\Big|$ and evaluate performance with MAE and Mean Squared Error (MSE) $\mathcal{R}
= \frac{1}{DH}\sum_{d=1}^{D}\sum_{h=1}^{H}\Big(y_{k_d+h}-\hat y_{k_d, h}\Big)^2$. All datasets are standardized using mean and standard deviation statistics computed from the respective training sets, and performance is reported in standardized units.

\paragraph{Favorita Sales Dataset.} The Favorita Sales Dataset~\citep{KaggleFavorita} is a large-scale retail forecasting dataset that contains daily sales from Corporación Favorita, a large grocery retailer located in Ecuador. The target time series are the total sales for a product family at a particular store on a given day. Future-known covariates are: (1) the total number of items in a product family that were being promoted at a store on a given day, and (2) whether the given day is a holiday. The number of promoted items is log-transformed, and the holiday indicator is binary encoded. In addition, we include day-of-week and day-of-month features as future-known covariates encoded with sine and cosine functions, yielding a total of six future-known covariates. As the Ecuadorian economy is highly dependent on oil, the daily oil price is included as a covariate for the past time steps. The oil price is standardized using mean and standard deviation statistics computed from the training set. Additionally, the dataset includes the product family, store number, city, state and type as well as a cluster variable, which groups similar stores, as categorical static covariates for each time series.

We use the preprocessed dataset from~\citet{shchurFevbenchRealisticBenchmark2025} and also use the same horizon of 28 days. We hold out a single evaluation window consisting of the last 28 days of all series as the test set and the preceding 28 days as the validation set. We fix the context length to 56 days. We train all models to minimize an MSE loss in log-transformed space as the primary forecasting loss.
As in the related Kaggle competition~\citep{KaggleFavorita}, we evaluate performance using the Root Mean Squared Logarithmic Error (RMSLE): $\mathcal{R}
= \sqrt{\frac{1}{DH}\sum_{d=1}^{D}\sum_{h=1}^{H}\Big(\log(1+y_{k_d+h})-\log(1+\hat y_{k_d, k_d+h})\Big)^2 }$. 

\paragraph{Bouc-Wen Hysteretic System.} The Bouc-Wen benchmark~\citep{schoukensThreeBenchmarksAddressing2017, boucwendataset} provides two noiseless datasets, originally intended for system identification: (a) a random phase multisine input dataset containing a steady state period of 8192 timesteps and (b) a sine-sweep input dataset in which the sine input signal increases frequency in a frequency band from 20 to 50 Hz. This dataset contains about 153,000 timesteps.

We use this benchmark as a controlled test bed for forecasting with exogenous inputs and define the following two forecasting tasks:
\begin{enumerate}
    \item \textbf{Multisine:} We employ a 70\%/10\%/20\% split on dataset (a) and extract samples using a stride (step size) of 1.
    \item \textbf{Transfer:} We train on the sinesweep dataset (b) of which we hold out the last 15\% as a validation set and evaluate on the multisine dataset (a). Samples are extracted using a stride of 24. This setting is designed to test generalization under a shift in the exogenous forcing process.

\end{enumerate}
For both tasks, we optimize the mean absolute error loss, use a context length of 288 steps, and a forecast horizon of 144 steps. Both datasets are standardized using mean and standard deviation statistics computed from the respective training sets, and performance is reported in standardized units.

\textbf{Dataset availability.}
We do not redistribute or re-upload any of the datasets used in this work. The electricity price forecasting (EPF) datasets~\citep{lagoForecastingDayaheadElectricity2021} are available from Zenodo\footnote{\url{https://zenodo.org/records/4624805}} the record lists the license as ``Other (Open)'', with source-specific terms applying to the original electricity-market data. The Favorita store-sales data were accessed through the \texttt{fev\_datasets} collection~\citep{shchurFevbenchRealisticBenchmark2025} on HuggingFace\footnote{\url{https://huggingface.co/datasets/autogluon/fev_datasets}}, which identifies the original source as Kaggle's \emph{Store Sales -- Time Series Forecasting} competition\footnote{\url{https://www.kaggle.com/competitions/store-sales-time-series-forecasting}}~\citep{KaggleFavorita}; the Hugging Face record is licensed as ``other'' and points users to the original source terms, while Kaggle permits non-commercial use under the competition rules. The Bouc-Wen hysteresis benchmark data~\citep{schoukensThreeBenchmarksAddressing2017, boucwendataset} are available from 4TU.ResearchData\footnote{\url{https://data.4tu.nl/articles/_/12967592}} under the Creative Commons Attribution-ShareAlike 4.0 International license (CC BY-SA 4.0). 

\subsection{Additional information on baselines and training methodology} \label{app:baselines+train_method}

We consider two groups of baselines: competitive forecasting architectures with exogenous-variable support, and related structured latent-dynamics baselines. For each baseline, we summarize the model class, the implementation used in our experiments, and any adaptations required to match our covariate setting.

\textbf{TimeXer.} The TimeXer model~\citep{wangTimeXerEmpoweringTransformers2024} is a Transformer-based architecture that is specifically designed for forecasting with exogenous variables. As the original implementation assumes that endogenous and exogenous sequences have the same length, it cannot directly accommodate future-known covariates beyond the historical context. We therefore use the implementation from~\citet{GCGNetGraphConsistentGenerative2025}, which modifies the exogenous-input embedding to support future-known covariates in the forecasting setting studied here. However, this version is not able to utilize past-only covariates without further modifications, which only affects the oil price variable in the Favorita task.

\textbf{TSMixer.} The Time-Series Mixer (TSMixer)~\citep{chenTSMixerAllMLPArchitecture2023} is an MLP-based architecture built around mixing operations along the time and feature dimensions.
We use the implementation from Neuralforecast~\citep{olivares2022library_neuralforecast} to run our experiments. TSMixer natively incorporates future-known, past-only and static covariates (in the TSMixerx variant).

\textbf{GCGNet.} The Graph-Consistent Generative Network (GCGNet)~\citep{GCGNetGraphConsistentGenerative2025} is a recently proposed graph-based time series forecasting model that is specifically designed for forecasting tasks with exogenous variables. We use the original implementation to run baseline experiments. 
On the Favorita dataset, GCGNet exhibited unstable (internal) variance estimates; we therefore clamp these values to a bounded interval in order to stabilize training on this dataset.

\textbf{TiDE.} Time-series Dense Encoder (TiDE)~\citep{dasLongtermForecastingTiDE2023} is an MLP-based encoder-decoder architecture. We use the implementation from Neuralforecast~\citep{olivares2022library_neuralforecast} to run our experiments.
TiDE natively incorporates future-known, past-only and static covariates.

\textbf{SKOLR.} Structured Koopman Operator Linear Recurrent Neural Network (SKOLR)~\citep{zhangSKOLRStructuredKoopman2025a} is a Koopman-based architecture that is structurally related to our approach. However, it is not able to incorporate future-known exogenous covariates. We use the original implementation and treat the past-only and the past portion of future-known covariates as additional channels of the input and run SKOLR in the non-channel-independent mode.

\textbf{DLinear.} DLinear~\citep{zengAreTransformersEffective2023} is a simple linear architecture that includes a seasonal decomposition. We use the implementation from DARTS~\citep{darts}, which natively incorporates past-only, future-known and static covariates.

\textbf{Koopman-Subnet.} Koopman-Subnet~\citep{iacobDeepIdentificationNonlinear2021} is a neural Koopman-with-input architecture originally developed for nonlinear system identification. It learns a lifted latent representation together with a generic MLP-based encoder and a state-dependent input map. We use the original implementation and adapt the inputs to our forecasting setting. Relative to \modelname{}, Koopman-Subnet uses a more generic MLP encoder and a state-dependent input map,

\textbf{CT-Subnet.} CT-Subnet~\citep{beintemaContinuoustimeIdentificationDynamic2022} is an architecture developed from a continuous-time system-identification perspective. We include it because it is structurally close to our setting in that it learns a latent state from observed history, but it differs from \modelname{} in several key respects: it uses a generic MLP encoder, models nonlinear latent dynamics rather than a shared linear latent transport operator, and is motivated by continuous-time identification rather than discrete-time exogenous forecasting. We evaluate CT-Subnet using its original implementation.

\textbf{MamKO.} MamKO~\citep{liMamKOMambabasedKoopman2024} is an adaptive Koopman-style architecture inspired by Mamba~\citep{guMambaLinearTimeSequence2024} and designed for modeling and control. Unlike \modelname{}, which keeps the latent transition fixed and uses the full history to reconstruct a forecast-start latent state, MamKO uses historical context to generate operator-related components while obtaining the initial latent state from only the last target value through an MLP. We include MamKO as a conceptually relevant baseline because it represents an alternative design in which context modulates the latent operator itself rather than improving latent-state estimation at rollout start. We evaluate MamKO using its original implementation. In the original MamKO formulation, the input at time $t$ drives the state transition from $t$ to $t+1$. 
We additionally evaluate the model with the complete exogenous input sequence shifted forward by one time step. This aligns the forecast covariates with the predicted timestamps and includes the final forecast covariate row in the rollout. Changes that noticeably exceed the reported standard deviation across seeds are improvements in EPEX BE MAE and MSE and EPEX FR MAE, and a further deterioration on the Bouc–Wen transfer task. The overall conclusions and discussions of MamKO remain qualitatively unchanged.
We do not evaluate shifted Subnet variants as these architectures support direct feedthrough from $\bu_t$ to $\by_t$.

\textbf{Baseline Implementations.} GCGNet~\citep{GCGNetGraphConsistentGenerative2025} and TimeXer~\citep{wangTimeXerEmpoweringTransformers2024} were implemented using the official GCGNet repository\footnote{\url{https://github.com/decisionintelligence/GCGNet}}, which is released under the MIT license. TiDE~\citep{dasLongtermForecastingTiDE2023} and TSMixer~\citep{chenTSMixerAllMLPArchitecture2023} were run using NeuralForecast\footnote{\url{https://github.com/nixtla/neuralforecast}}~\citep{olivares2022library_neuralforecast}, which is released under the Apache License 2.0. DLinear~\citep{zengAreTransformersEffective2023} was run using Darts\footnote{\url{https://github.com/unit8co/darts/tree/master}}~\citep{darts}, which is also released under the Apache License 2.0. SKOLR~\citep{zhangSKOLRStructuredKoopman2025a} was run from its public repository\footnote{\url{https://github.com/networkslab/SKOLR}}, which is released under the MIT license. The CT-Subnet~\citep{beintemaContinuoustimeIdentificationDynamic2022} and Koopman-Subnet~\citep{iacobDeepIdentificationNonlinear2021} baselines were implemented using the deepSI toolbox\footnote{\url{https://github.com/MaartenSchoukens/deepSI}}, which is released under the BSD 3-Clause license. MamKO~\citep{liMamKOMambabasedKoopman2024} was implemented based on the CC BY 4.0-licensed OpenReview submission and its accompanying supplementary material\footnote{\url{https://openreview.net/forum?id=hNjCVVm0EQ}}. 

\begin{table}[ht]
    \centering
    \caption{Usage of past-only, future-known and static exogenous features in baselines. (\cmark) denotes models that do not natively incorporate static or past features. In this case, we provide an \emph{optional} mechanism for static features that repeats the embedding and concatenates it to the inputs at each time step. For past features, we add their sequence as an additional channel to the target, while keeping a single output channel, if this is possible without major modifications.}
    \begin{tabular}{lccc}
    \toprule
     \textbf{Baseline}    &  \textbf{Past} & \textbf{Future} & \textbf{Static}\\ \midrule
      TimeXer   &  \xmark & \cmark & (\cmark)     \\ 
      TSMixer   &  \cmark & \cmark & \cmark\\ 
      GCGNet   &  \xmark & \cmark & (\cmark)    \\ 
      TiDE   &  \cmark & \cmark & \cmark  \\ 
      DLinear   &  \cmark & \cmark & \cmark  \\ 
      SKOLR   &  (\cmark) & \xmark & (\cmark)  \\ \midrule
      Koopman-Subnet   &  (\cmark) & \cmark & (\cmark)     \\ 
      CT-Subnet   &  (\cmark) & \cmark & (\cmark)     \\ 
      MamKO  &  (\cmark) & \cmark & (\cmark)     \\ \bottomrule
    \end{tabular}
    \label{tab:exo}
\end{table}

\paragraph{Categorical static embedding.} To allow the use of categorical static variables, we adopt the same learnable static embedding scheme for all baselines. For the baselines that do not natively incorporate static variables (TimeXer, GCGNet, SKOLR),  we provide an optional mechanism that repeats the embedding and concatenates it to the inputs at each time step. We treat the inclusion of static information as a binary hyperparameter during tuning, allowing these models to leverage the static features if beneficial without forcing a non-native design choice.

\paragraph{Instance stationarization.}  Differences in instance stationarization can substantially influence forecasting accuracy~\citep{kimReversibleInstanceNormalization2021}. To isolate the contribution of each method, we fix the instance stationarization protocol across all baselines for a given dataset, choosing the setting that performed best across baseline models. In particular, we apply instance stationarization to both endogenous and exogenous inputs for the EPF-BE and EPF-FR datasets, while applying it only to the endogenous input signal for the remaining EPF datasets and Favorita. We do not apply any instance stationarization to the Bouc-Wen tasks. 
An exception is Koopman-Subnet, which occasionally diverged at test time, without additional stationarization,in some settings. For this model, we also apply instance stationarization to exogenous inputs on EPF-DE, and to both endogenous and exogenous inputs on the Bouc--Wen transfer task.

\paragraph{Hyperparameter optimization.} 
For all baseline models, we define the hyperparameter search spaces based on the ranges reported in the original papers and list them for the EPF datasets in \cref{tab:hpo}. We use a random search with a fixed budget of 30 trials per model and dataset. Each configuration is evaluated on the validation set using early stopping with a patience of five epochs, and the configuration with the best validation performance is selected as the final model setting.  The final baseline metrics are reported as averages over five independent runs with different random seeds. For the ablation experiments, we use a smaller random-search budget of 8 trials centered around the best configuration of the base model.

\begin{table}[ht!]
    \tiny
    \centering
    \caption{Hyperparameter optimization ranges for \modelname{} and baselines on all datasets.}
    \label{tab:hpo}
    {\setlength{\tabcolsep}{6pt}
    \begin{tabular}{lccc}
    \toprule
      \textbf{Hyperparameter}   & \textbf{EPF} & \textbf{Favorita} & \textbf{Bouc-Wen}  \\ \midrule
        \textbf{\modelname{}}  &   \\
        ~~batch size  & \{64, 128, 256\} & \{256, 512\} & \{64, 128, 256\}\\
        ~~learning rate & \{$10^{-5}$, $5 \cdot10^{-5}$, $10^{-4}$\} & \{$5 \cdot10^{-5}$, $10^{-4}$\}  & \{$10^{-5}$, $5 \cdot10^{-5}$, $10^{-4}$\}\\
        ~~blocks & \{2\} & \{2\} & \{2\} \\
        ~~dynamic dim.  & \{128, 256\} & \{128, 256\} & \{128, 256\}\\
        ~~hidden layers  & \{2\} & \{2\} & \{2\} \\
        ~~dropout & \{0.05, 0.15\} & \{0.05, 0.15\} & \{0.05, 0.15\} \\
        ~~patch length & \{24\} & \{4, 7\} & \{4, 8\} \\ \midrule
        \textbf{TimeXer}  &   \\
        ~~batch size  & \{16, 32, 64\} & \{128, 256, 512\} & \{16, 32, 64\} \\
        ~~learning rate & \{$10^{-2}$, $10^{-3}$, $10^{-4}$, $10^{-5}$\} & \{$10^{-2}$, $10^{-3}$, $10^{-4}$, $10^{-5}$\} & \{$10^{-2}$, $10^{-3}$, $10^{-4}$, $10^{-5}$\}\\
        ~~feedforward dim. factor  & \{1, 2, 4, 8\}  & \{1, 2, 4, 8\} & \{1, 2, 4, 8\} \\
        ~~model dim.  & \{128, 256, 512\} & \{128, 256, 512\} & \{128, 256, 512\} \\
        ~~encoder layer & \{1, 2, 3\} & \{1, 2, 3\} & \{1, 2, 3\} \\
        ~~number of heads & \{4, 8\} & \{4, 8\} & \{4, 8\} \\
        ~~patch length  & \{24\} & \{4, 7\} & \{4, 12, 24\}\\ \midrule
        \textbf{TSMixer} &   \\
        ~~batch size  & \{16, 32, 64\} & \{64, 128, 256\} & \{16, 32, 64\}\\
        ~~learning rate & \{$10^{-2}$, $10^{-3}$, $10^{-4}$, $10^{-5}$\} & \{$10^{-2}$, $10^{-3}$, $10^{-4}$, $10^{-5}$\}  & \{$10^{-2}$, $10^{-3}$, $10^{-4}$, $10^{-5}$\}\\
        ~~blocks  & \{1, 2, 4, 6, 8\} & \{1, 2, 4, 6, 8\} & \{1, 2, 4, 6, 8\} \\
        ~~dropout  & \{0, 0.1, 0.3, 0.5, 0.9\} & \{0, 0.1, 0.3, 0.5, 0.9\} & \{0, 0.1, 0.3, 0.5, 0.9\} \\
        ~~feedforward dim.  & \{16, 32, 64, 128, 256\} & \{16, 32, 64, 128, 256\} & \{16, 32, 64, 128, 256\} \\\midrule
        \textbf{GCGNet}  &   \\
        ~~batch size  & \{16, 32, 64\} & \{128, 256, 512\} & \{16, 32, 64\} \\
        ~~learning rate & \{$10^{-2}$, $10^{-3}$, $10^{-4}$, $10^{-5}$\} & \{$10^{-2}$, $10^{-3}$, $10^{-4}$, $10^{-5}$\} & \{$10^{-2}$, $10^{-3}$, $10^{-4}$, $10^{-5}$\}\\
        ~~feedforward dim.  & \{64, 128, 256, 512\} & \{64, 128, 256, 512\}  & \{64, 128, 256, 512\} \\
        ~~model dim. & \{64, 128, 256, 512\} & \{64, 128, 256, 512\} & \{64, 128, 256, 512\}  \\
        ~~heads & \{4\} & \{4\} & \{4\} \\\midrule
        \textbf{TiDE}  &   \\
        ~~batch size & \{512\} & \{512\} & \{512\} \\
        ~~learning rate & \{$10^{-2}$, $10^{-3}$, $10^{-4}$, $10^{-5}$\} & \{$10^{-2}$, $10^{-3}$, $10^{-4}$, $10^{-5}$\} & \{$10^{-2}$, $10^{-3}$, $10^{-4}$, $10^{-5}$\}\\
        ~~hidden size  & \{256, 512, 1024\} & \{256, 512, 1024\} & \{256, 512, 1024\} \\
        ~~encoder layer  & \{1, 2, 3\} & \{1, 2, 3\}& \{1, 2, 3\} \\
        ~~decoder layer  & \{1, 2, 3\} & \{1, 2, 3\}& \{1, 2, 3\}\\
        ~~decoder output dim. & \{4, 8, 16, 32\} & \{4, 8, 16, 32\} & \{4, 8, 16, 32\} \\
        ~~temporal decoder dim. & \{32, 64, 128\} & \{32, 64, 128\} & \{32, 64, 128\} \\
        ~~dropout  & \{0, 0.1, 0.2, 0.3, 0.5\} & \{0, 0.1, 0.2, 0.3, 0.5\} & \{0, 0.1, 0.2, 0.3, 0.5\} \\
        ~~layerNorm & \{True, False\} & \{True, False\} & \{True, False\} \\\midrule
        \textbf{DLinear}  &   \\
        ~~batch size  & \{32, 64, 128, 256, 512\} & \{64, 128, 256, 512\}  & \{32, 64, 128, 256, 512\}\\
        ~~learning rate & \{$10^{-2}$, $10^{-3}$, $10^{-4}$, $10^{-5}$\} & \{$10^{-2}$, $10^{-3}$, $10^{-4}$, $10^{-5}$\} & \{$10^{-2}$, $10^{-3}$, $10^{-4}$, $10^{-5}$\}\\
        ~~kernel size  & \{24\} & \{7\} & \{4, 12, 24, 48\} \\
        ~~constant initialization & \{True, False\} & \{True, False\} & \{True, False\} \\\midrule
        \textbf{SKOLR}  &   \\
        ~~batch size  & \{16, 32, 64\} & \{128, 256, 512\} & \{16, 32, 64\} \\
        ~~learning rate & \{$10^{-2}$, $10^{-3}$, $10^{-4}$, $10^{-5}$\} & \{$10^{-2}$, $10^{-3}$, $10^{-4}$, $10^{-5}$\} & \{$10^{-2}$, $10^{-3}$, $10^{-4}$, $10^{-5}$\}\\
        ~~blocks  & \{2, 3, 4, 8\} & \{2, 3, 4, 8\} & \{2, 3, 4, 8\} \\
        ~~hidden layers & \{ 1, 2, 3 \} & \{ 1, 2, 3 \}& \{ 1, 2, 3 \} \\
        ~~dynamic dim. & \{ 128, 256, 512 \}  & \{ 128, 256, 512 \} & \{ 128, 256, 512 \} \\
        ~~dropout & \{ 0.0, 0.05, 0.1, 0.2 \} & \{ 0.0, 0.05, 0.1, 0.2 \}& \{ 0.0, 0.05, 0.1, 0.2 \} \\
        ~~patch length & \{12, 24\} & \{4, 7\} &  \{4, 8\}  \\ \midrule
        \textbf{CT-Subnet}  &   \\
        ~~batch size  & \{64, 128, 256\} & \{128, 256, 512\} & \{64, 128, 256\} \\
        ~~learning rate & \{$10^{-2}$, $10^{-3}$, $10^{-4}$\} & \{$10^{-2}$, $10^{-3}$, $10^{-4}$\} & \{$10^{-2}$, $10^{-3}$, $10^{-4}$\}\\
        ~~latent dim.  & \{16, 32, 64, 128, 256\} & \{16, 32, 64, 128, 256\} & \{16, 32, 64, 128, 256\} \\
        ~~feedthrough & \{True, False\} & \{True, False\}& \{True, False\} \\
        \midrule
        \textbf{Koop-Subnet}  &   \\
        ~~batch size  & \{64, 128, 256\} & \{128, 256, 512\} & \{64, 128, 256\} \\
        ~~learning rate & \{$10^{-2}$, $10^{-3}$, $10^{-4}$\} & \{$10^{-2}$, $10^{-3}$, $10^{-4}$\} & \{$10^{-2}$, $10^{-3}$, $10^{-4}$\}\\
        ~~latent dim.  & \{16, 32, 64, 128, 256\} & \{16, 32, 64, 128, 256\} & \{16, 32, 64, 128, 256\} \\
        ~~feedthrough & \{True, False\} & \{True, False\}& \{True, False\} \\
        ~~u-dependent input map & \{True, False\} & \{True, False\}& \{True, False\} \\ \midrule
        \textbf{MamKO}  &   \\
        ~~batch size  & \{64, 128, 256\} & \{256, 512\} & \{64, 128, 256\} \\
        ~~learning rate & \{$10^{-2}$, $10^{-3}$, $10^{-4}$\} & \{$10^{-2}$, $10^{-3}$, $10^{-4}$\} & \{$10^{-2}$, $10^{-3}$, $10^{-4}$\}\\
        ~~kernel size & \{2, 24, 48\}  & \{7, 14, 21\} & \{18, 36, 72\} \\
        ~~delta (factor of latent) & \{0.25, 0.5, 1\} & \{0.25, 0.5, 1\} & \{0.25, 0.5, 1\} \\
        ~~latent dim. & \{ 16, 32, 64, 128\} & \{ 16, 32, 64, 128\} & \{ 16, 32, 64, 128\}\\
        ~~hidden dim. & \{ 64, 128, 256 \} & \{ 64, 128, 256 \} & \{ 64, 128, 256 \} \\
        ~~learning rate scheduler  & \{True, False\} & \{True, False\}& \{True, False\} \\
        
       \bottomrule
    \end{tabular}}
\end{table}

\subsection{Computational resources} \label{app:comp_resources}
Experiments were conducted on multiple comparable compute systems. For concreteness, we report one representative configuration used for evaluation: a single node of a high-performance computing system with four NVIDIA A100 GPUs (\qty{40}{\giga\byte}), an Intel Xeon Platinum 8368 processor, and \qty{512}{\giga\byte} RAM. Note that each individual training run used a single GPU. The remaining runs were performed on machines with comparable hardware configurations, which is not expected to materially affect the reported results. The computational efficiency runs were all made on the same hardware for comparability.
Overall, the experiments required approximately 1,000 GPU hours.

\section{Performance of Koopman/system-identification baselines across benchmarks}

We observe that Koopman-Subnet, CT-Subnet, and MamKO perform most favorably in settings that closely resemble classical system identification, with compact dynamics and directly observed forcing signals. This is visible in the Bouc--Wen experiments, where the methods perform strongly on Multisine, but degrade substantially under the distribution shift of the Transfer task and when future exogenous inputs are perturbed.

To test whether this trend extends beyond the original benchmarks, we additionally evaluate on the real-world UCI Bike Sharing dataset~\citep{fanaee-tBikeSharing2013} and a synthetic Duffing oscillator driven by band-limited Gaussian forcing.

\begin{table}[ht!]
    \centering
    \caption{Results of Koopman/system-identification baselines on two additional dataset. Best results are highlighted in \textbf{bold}.}
    \begin{tabular}{lcccc}
    \toprule
     & \multicolumn{2}{c}{\textbf{Bike}} & \multicolumn{2}{c}{\textbf{Duffing}}  \\
     \cmidrule(lr){2-3}\cmidrule(lr){4-5}
    \textbf{Model} & \textbf{MAE} & \textbf{MSE} & \textbf{MAE} & \textbf{MSE} \\
    \midrule
    SLiDE & \textbf{0.230} & \textbf{0.129} & 0.204 & 0.089 \\
    TimeXer & 0.301 & 0.239 & 0.218 & 0.091\\
    GCGNet & 0.259 & 0.174 & 0.386 & 0.283 \\
    CT-Subnet & 0.267 & 0.162 & \textbf{0.119} & \textbf{0.029} \\
    Koop.-Subnet & 0.315 & 0.211 & 0.383 & 0.248  \\
    MamKO  & 0.327 & 0.262 & 0.641 & 0.665 \\
    \bottomrule
    \end{tabular}
    \label{tab:bikeduff}
\end{table}

The additional results shown in \cref{tab:bikeduff} follow this pattern. On Bike Sharing, the system-identification methods show no systematic advantage. CT-Subnet remains competitive, while SLiDE performs best and Koopman-Subnet and MamKO trail the forecasting baselines. On Duffing, which more closely matches a controlled system-identification setting, CT-Subnet performs strongly. At the same time, the weaker Duffing results of Koopman-Subnet and MamKO show that this is not a uniform advantage of all system-identification or control methods.

\section{Statistical evaluation} \label{app:stats}

\textbf{Statistical Test Setup.} We assess statistical significance using one-sided pairwise tests. For all datasets except Favorita, we apply the one-sided Diebold–Mariano (DM) test on the time series of error differences per sample averaged over seeds. We set the truncation of the variance estimate to the horizon divided by the step size between test samples. We implemented the DM test in Python following the implementation of the R forecast package~\citep{hyndmanforecastpackage}. As each Favorita test sample corresponds to a different product group predicted over the same time range, we instead use a one-sided Wilcoxon signed-rank test on per-product-group error differences, again averaged over seeds. We deem results significant at $p < 0.05$.

\textbf{Bootstrap Confidence Intervals.} In addition, we report $95\%$ confidence intervals of the loss differential (again averaged over the five seeds) obtained via stationary bootstrap applying automatic block-length selection depending on autocorrelation in \cref{tab:CI}. The bootstrap confidence intervals show that \modelname{}’s advantage is broad but not uniform, with the strongest and most statistically stable gains appearing on EPEX-DE and especially on the Bouc–Wen transfer task.

\begin{table*}[ht]
    \centering
    \caption{Bootstrap confidence intervals of the performance differences between \modelname{} and baselines. The reported numbers represent the lower and upper bounds of the 95\% confidence intervals.}
    \label{tab:CI}
    \scriptsize
    {\setlength{\tabcolsep}{0.7pt} %
    \begin{tabular}{lccccccccccccccc}
    \toprule
     & \multicolumn{10}{c}{\textbf{EPF Benchmark}} & \textbf{Favorita} & \multicolumn{4}{c}{\textbf{Bouc-Wen System}} \\\cmidrule(lr){2-11}\cmidrule(lr){12-12}\cmidrule(lr){13-16}
       \textbf{Model}  & \multicolumn{2}{c}{\textbf{NP}} & \multicolumn{2}{c}{\textbf{PJM}} & \multicolumn{2}{c}{\textbf{EPEX BE}} & \multicolumn{2}{c}{\textbf{EPEX FR}} & \multicolumn{2}{c}{\textbf{EPEX DE}} & \textbf{Sales} & \multicolumn{2}{c}{\textbf{Multisine}} & \multicolumn{2}{c}{\textbf{Transfer}} \\ 
           \cmidrule(lr){2-3} \cmidrule(lr){4-5} \cmidrule(lr){6-7} \cmidrule(lr){8-9} \cmidrule(lr){10-11} 
           \cmidrule(lr){12-12} \cmidrule(lr){13-14}\cmidrule(lr){15-16}
      & $\Delta$MAE & $\Delta$MSE & $\Delta$MAE & $\Delta$MSE& $\Delta$MAE & $\Delta$MSE&  $\Delta$MAE & $\Delta$MSE& 
       $\Delta$MAE& 
      $\Delta$MSE& 
      $\Delta$MSLE & $\Delta$MAE & $\Delta$MSE& $\Delta$MAE & $\Delta$MSE \\\midrule
        
      \modelname{} & - & - & - & - &  - & - &  - & - &  - & - &  - & - &  - & - &  - \\ \midrule
  TimeXer & \ci{$0.004$}{$0.017$} & \ci{$0.008$}{$0.033$} & \ci{$-0.007$}{$0.001$} & \ci{$-0.010$}{$0.001$} &  \ci{$-0.006$}{$0.003$} & \ci{$-0.006$}{$0.007$} &  \ci{$-0.007$}{$0.005$} &  \ci{$-0.003$}{$0.053$} & \ci{$0.005$}{$0.021$}   & \ci{$0.012$}{$0.053$} &  \ci{$0.018$}{$0.060$} &    \ci{$0.069$}{$0.082$} & \ci{$0.025$}{$0.033$} & \ci{$0.099$}{$0.145$} & \ci{$0.106$}{$0.158$} \\ \midrule
  TSMixer & \ci{$0.003$}{$0.016$} & \ci{$0.001$}{$0.039$} &  \ci{$-0.003$}{$0.005$} & \ci{$-0.006$}{$0.002$} & \ci{$-0.001$}{$0.009$} & \ci{$-0.083$}{$0.006$} &  \ci{$-0.001$}{$0.057$} &  \ci{$-0.026$}{$0.028$} &  \ci{$0.006$}{$0.034$}  & \ci{$0.006$}{$0.076$} &   \ci{$-0.003$}{$0.009$} &    \ci{$0.002$}{$0.012$} & \ci{$0.000$}{$0.003$} & \ci{$0.224$}{$0.265$} & \ci{$0.312$}{$0.368$} \\ \midrule
   GCGNet & \ci{$0.005$}{$0.015$} & \ci{$-0.001$}{$0.024$} & \ci{$-0.003$}{$0.003$} & \ci{$-0.002$}{$0.004$} &  \ci{$0.014$}{$0.029$} & \ci{$-0.048$}{$0.033$} & \ci{$0.003$}{$0.049$} &  \ci{$-0.048$}{$0.019$} &  \ci{$0.011$}{$0.031$}    & \ci{$0.007$}{$0.048$} &  \ci{$0.030$}{$0.088$} &   \ci{$0.129$}{$0.141$} & \ci{$0.071$}{$0.078$} & \ci{$0.127$}{$0.175$} & \ci{$0.153$}{$0.210$} \\ \midrule
  TiDE & \ci{$0.026$}{$0.042$} & \ci{$0.021$}{$0.046$} & \ci{$0.002$}{$0.017$} & \ci{$0.000$}{$0.048$} &  \ci{$0.016$}{$0.034$} & \ci{$-0.056$}{$0.030$} &  \ci{$0.004$}{$0.082$} &    \ci{$-0.025$}{$0.081$} &  \ci{$0.017$}{$0.046$}  & \ci{$0.015$}{$0.071$} &  \ci{$0.000$}{$0.009$} & \ci{$0.050$}{$0.066$} & \ci{$0.016$}{$0.022$} & \ci{$0.164$}{$0.212$} & \ci{$0.214$}{$0.275$} \\ \midrule
  DLinear & \ci{$0.028$}{$0.050$} & \ci{$0.030$}{$0.061$} & \ci{$0.002$}{$0.017$} & \ci{$0.001$}{$0.023$} & \ci{$0.010$}{$0.049$} & \ci{$-0.071$}{$0.059$} &  \ci{$0.003$}{$0.076$} &   \ci{$-0.008$}{$0.094$} &  \ci{$0.021$}{$0.069$}  & \ci{$0.024$}{$0.125$} &  \ci{$0.032$}{$0.088$} &    \ci{$0.667$}{$0.749$} & \ci{$0.857$}{$1.050$} & \ci{$1.025$}{$1.089$} & \ci{$2.674$}{$2.887$} \\ \midrule
  SKOLR & \ci{$0.041$}{$0.069$} & \ci{$0.048$}{$0.094$} & \ci{$0.019$}{$0.040$} & \ci{$0.014$}{$0.064$} & \ci{$0.016$}{$0.034$} & \ci{$0.007$}{$0.093$} &  \ci{$0.020$}{$0.047$} &  \ci{$0.011$}{$0.098$} & \ci{$0.123$}{$0.168$}    & \ci{$0.180$}{$0.331$} &  \ci{$0.024$}{$0.086$} &     \ci{$0.629$}{$0.710$} & \ci{$0.761$}{$0.940$} & \ci{$0.463$}{$0.507$} & \ci{$0.761$}{$0.821$} \\ 
  \midrule
Koop.-Subnet & \ci{$0.030$}{$0.053$} & \ci{$0.023$}{$0.081$} & \ci{$0.016$}{$0.034$} & \ci{$0.008$}{$0.056$} & \ci{$0.012$}{$0.026$} & \ci{$-0.039$}{$0.018$} & \ci{$0.005$}{$0.066$} & \ci{$-0.009$}{$0.051$} & \ci{$0.035$}{$0.061$} & \ci{$0.047$}{$0.105$} & \ci{$0.012$}{$0.022$} & \ci{$-0.028$}{$-0.013$} & \ci{$-0.007$}{$-0.003$}  & \ci{$0.402$}{$0.449$} & \ci{$0.628$}{$0.721$}
\\ \midrule
  CT-Subnet & \ci{$0.020$}{$0.034$} & \ci{$0.010$}{$0.029$} & \ci{$0.012$}{$0.027$} & \ci{$0.006$}{$0.040$} & \ci{$0.019$}{$0.042$} & \ci{$0.001$}{$0.043$} & \ci{$0.017$}{$0.077$} & \ci{$0.007$}{$0.119$} & \ci{$0.038$}{$0.071$} & \ci{$0.034$}{$0.107$} & \ci{$0.004$}{$0.012$} & \ci{$-0.021$}{$-0.008$} &  \ci{$-0.004$}{$0.000$} & \ci{$0.447$}{$0.496$} & \ci{$0.700$}{$0.792$} \\ \midrule
 MamKO & \ci{$0.027$}{$0.045$} & \ci{$0.032$}{$0.064$} & \ci{$0.010$}{$0.021$} & \ci{$0.004$}{$0.031$} & \ci{$0.019$}{$0.031$} & \ci{$-0.014$}{$0.023$} & \ci{$0.011$}{$0.039$} & \ci{$-0.121$}{$0.080$} & \ci{$0.043$}{$0.083$} & \ci{$0.052$}{$0.152$} & \ci{$0.041$}{$0.089$} & \ci{$0.356$}{$0.402$} & \ci{$0.296$}{$0.360$}  & \ci{$0.607$}{$0.689$} & \ci{$1.391$}{$1.787$} \\
     
      \bottomrule
    \end{tabular}}
\end{table*}

\textbf{Variability Across Random Seeds.} To assess the stability of the evaluated methods, we report the standard deviation of performance across five random seeds for each baseline–dataset combination in Table~\cref{tab:seeds}.

\begin{table*}[ht]
    \centering
    \caption{Standard deviation of performance across five random seeds for each baseline and dataset. We report the population standard deviation, computed over the five runs. Lower values indicate greater stability across random seeds.}
    \label{tab:seeds}
    \tiny
    {\setlength{\tabcolsep}{0.25pt} %
    \begin{tabular}{lccccccccccccccc}
    \toprule
     & \multicolumn{10}{c}{\textbf{EPF Benchmark}} & \textbf{Favorita} & \multicolumn{4}{c}{\textbf{Bouc-Wen System}} \\\cmidrule(lr){2-11}\cmidrule(lr){12-12}\cmidrule(lr){13-16}
       \textbf{Model}  & \multicolumn{2}{c}{\textbf{NP}} & \multicolumn{2}{c}{\textbf{PJM}} & \multicolumn{2}{c}{\textbf{EPEX BE}} & \multicolumn{2}{c}{\textbf{EPEX FR}} & \multicolumn{2}{c}{\textbf{EPEX DE}} & \textbf{Sales} & \multicolumn{2}{c}{\textbf{Multisine}} & \multicolumn{2}{c}{\textbf{Transfer}} \\ 
           \cmidrule(lr){2-3} \cmidrule(lr){4-5} \cmidrule(lr){6-7} \cmidrule(lr){8-9} \cmidrule(lr){10-11} 
           \cmidrule(lr){12-12} \cmidrule(lr){13-14}\cmidrule(lr){15-16}
      & MAE & MSE & MAE & MSE & MAE & MSE &  MAE & MSE & MAE & MSE & RMSLE & MAE & MSE & MAE & MSE \\\midrule 
\modelname{} & \ci{\pm0.002}{0.210} & \ci{\pm0.003}{0.167} & \ci{\pm0.003}{0.157} & \ci{\pm0.004}{0.064} & \ci{\pm0.001}{0.231} & \ci{\pm0.003}{0.367} & \ci{\pm0.003}{0.169} & \ci{\pm0.003}{0.329} & \ci{\pm0.002}{0.265} & \ci{\pm0.003}{0.182} & \ci{\pm0.002}{0.419} & \ci{\pm0.001}{0.085} & \ci{\pm0.000}{0.012} & \ci{\pm0.010}{0.296} & \ci{\pm0.010}{0.139} \\
TimeXer & \ci{\pm0.002}{0.220} & \ci{\pm0.005}{0.186} & \ci{\pm0.001}{0.154} & \ci{\pm0.001}{0.060} & \ci{\pm0.003}{0.230} & \ci{\pm0.007}{0.367} & \ci{\pm0.001}{0.169} & \ci{\pm0.006}{0.347} & \ci{\pm0.002}{0.278} & \ci{\pm0.003}{0.213} & \ci{\pm0.005}{0.461}  & \ci{\pm0.001}{0.160} & \ci{\pm0.001}{0.040} & \ci{\pm0.004}{0.419} & \ci{\pm0.004}{0.273} \\
TSMixer & \ci{\pm0.004}{0.219} & \ci{\pm0.004}{0.184} & \ci{\pm0.002}{0.158} & \ci{\pm0.001}{0.063} & \ci{\pm0.004}{0.235} & \ci{\pm0.014}{0.335} & \ci{\pm0.005}{0.189} & \ci{\pm0.017}{0.331} & \ci{\pm0.005}{0.283} & \ci{\pm0.007}{0.218} & \ci{\pm0.002}{0.422} & \ci{\pm0.004}{0.092} & \ci{\pm0.001}{0.013} & \ci{\pm0.092}{0.541} & \ci{\pm0.169}{0.480} \\
GCGNet & \ci{\pm0.006}{0.220} & \ci{\pm0.011}{0.177} & \ci{\pm0.003}{0.157} & \ci{\pm0.003}{0.065} & \ci{\pm0.014}{0.252} & \ci{\pm0.021}{0.367} & \ci{\pm0.004}{0.189} & \ci{\pm0.006}{0.320} & \ci{\pm0.006}{0.285} & \ci{\pm0.012}{0.206} & \ci{\pm0.005}{0.482} & \ci{\pm0.086}{0.220} & \ci{\pm0.075}{0.086} & \ci{\pm0.027}{0.448} & \ci{\pm0.040}{0.322} \\
TiDE & \ci{\pm0.006}{0.243} & \ci{\pm0.005}{0.200} & \ci{\pm0.004}{0.165} & \ci{\pm0.009}{0.081} & \ci{\pm0.006}{0.256} & \ci{\pm0.015}{0.362} & \ci{\pm0.001}{0.200} & \ci{\pm0.010}{0.353} & \ci{\pm0.002}{0.295} & \ci{\pm0.001}{0.220} & \ci{\pm0.003}{0.424} & \ci{\pm0.003}{0.143} & \ci{\pm0.001}{0.031} & \ci{\pm0.094}{0.485} & \ci{\pm0.158}{0.385} \\
DLinear & \ci{\pm0.002}{0.249} & \ci{\pm0.003}{0.211} & \ci{\pm0.000}{0.166} & \ci{\pm0.000}{0.074} & \ci{\pm0.006}{0.258} & \ci{\pm0.011}{0.371} & \ci{\pm0.001}{0.198} & \ci{\pm0.004}{0.363} & \ci{\pm0.001}{0.307} & \ci{\pm0.001}{0.248} & \ci{\pm0.001}{0.483} & \ci{\pm0.011}{0.793} & \ci{\pm0.031}{0.966} & \ci{\pm0.204}{1.354} & \ci{\pm0.852}{2.919} \\
SKOLR & \ci{\pm0.001}{0.264} & \ci{\pm0.001}{0.237} & \ci{\pm0.001}{0.185} & \ci{\pm0.003}{0.097} & \ci{\pm0.002}{0.256} & \ci{\pm0.004}{0.405} & \ci{\pm0.003}{0.201} & \ci{\pm0.007}{0.372} & \ci{\pm0.003}{0.410} & \ci{\pm0.006}{0.432} & \ci{\pm0.007}{0.478} & \ci{\pm0.008}{0.754} & \ci{\pm0.024}{0.862} & \ci{\pm0.005}{0.782} & \ci{\pm0.013}{0.929} \\
K.-Subnet & \ci{\pm0.007}{0.251} & \ci{\pm0.010}{0.214} & \ci{\pm0.003}{0.181} & \ci{\pm0.002}{0.089} & \ci{\pm0.003}{0.250} & \ci{\pm0.004}{0.360} & \ci{\pm0.004}{0.196} & \ci{\pm0.011}{0.345} & \ci{\pm0.004}{0.312} & \ci{\pm0.008}{0.257} & \ci{\pm0.007}{0.438} & \ci{\pm0.005}{0.064} & \ci{\pm0.001}{0.007} & \ci{\pm0.040}{0.721} & \ci{\pm0.088}{0.815} \\
CT-Subnet & \ci{\pm0.002}{0.237} & \ci{\pm0.007}{0.185} & \ci{\pm0.002}{0.175} & \ci{\pm0.004}{0.082} & \ci{\pm0.008}{0.260} & \ci{\pm0.017}{0.391} & \ci{\pm0.003}{0.208} & \ci{\pm0.007}{0.375} & \ci{\pm0.003}{0.318} & \ci{\pm0.004}{0.247} & \ci{\pm0.003}{0.428} & \ci{\pm0.007}{0.070} & \ci{\pm0.002}{0.010} & \ci{\pm0.002}{0.767} & \ci{\pm0.005}{0.885} \\
MamKO & \ci{\pm0.003}{0.245} & \ci{\pm0.005}{0.213} & \ci{\pm0.002}{0.172} & \ci{\pm0.002}{0.079} & \ci{\pm0.003}{0.256} & \ci{\pm0.005}{0.374} & \ci{\pm0.001}{0.192} & \ci{\pm0.012}{0.324} & \ci{\pm0.001}{0.325} & \ci{\pm0.005}{0.276} & \ci{\pm0.011}{0.489} & \ci{\pm0.015}{0.464} & \ci{\pm0.020}{0.340} & \ci{\pm0.135}{0.944} & \ci{\pm0.764}{1.722} \\
      \bottomrule
    \end{tabular}}
\end{table*}

\section{Extended related work}
\label{app:extended_related_work}

\textbf{Deep Learning for forecasting with exogenous variables.}
Deep learning has become an important paradigm for time series forecasting, and explicitly incorporating exogenous variables to capture external drivers of temporal dynamics into neural forecasters remains an active research area.
DeepAR represents an early recurrent approach that conditions probabilistic forecasts on past observations together with covariates~\citep{salinasDeepARProbabilisticForecasting2020}. 
Subsequent work has adopted transformer-based architectures for forecasting with exogenous information. The Temporal Fusion Transformer~\citep{limTemporalFusionTransformers2021} introduced attention and gating mechanisms together with explicit handling of static and time-varying covariates. The TimeXer architecture~\citep{wangTimeXerEmpoweringTransformers2024} further explored transformer-based forecasting and includes covariates with a cross-attention mechanism.
A parallel line of work explored simpler feedforward designs that remain competitive for forecasting with exogenous variables. The Time-series Dense Encoder (TiDE)~\citep{dasLongtermForecastingTiDE2023} adopts a lightweight encoder–decoder formulation. TSMixer~\citep{chenTSMixerAllMLPArchitecture2023} is another all-MLP architecture that uses mixing operations across time and channels. N-BEATSx~\citep{olivaresNeuralBasisExpansion2023} extends the N-BEATS architecture~\citep{oreshkinNBEATSNeuralNetwork2021} to forecasting with exogenous covariates. 
Recently, GCGNet~\citep{GCGNetGraphConsistentGenerative2025} has been proposed as a generative, graph-based model that explicitly targets the future-known covariate setting. In contrast to attention-, graph-, and MLP-based approaches that primarily rely on flexible feature fusion, \modelname{} imposes a constrained input-driven latent dynamics model, which we show improves accuracy and efficiency while promoting generalization under shifted input processes.

\textbf{Koopman with inputs and controls.} The Koopman operator is extensively used to describe dynamical systems. Some works within this community consider additional inputs, such as (Extended) Dynamic Mode Decomposition with control (DMDc/EDMDc)~\citep{proctor2016dynamic, kordaLinearPredictorsNonlinear2018} and Koopman-with-input formulations~\citep{proctorGeneralizingKoopmanTheory2018, kaiser2021koopman}. 
These methods were developed primarily for system identification and control, where inputs are typically treated as observed, externally specified control or forcing signals and, in control applications, as directly manipulable. In contrast, forecasting applications involve \emph{exogenous variables} that act as external drivers rather than controls and are often partially observed. 
Beyond classical approaches with hand-chosen lifting functions, a substantial body of work has studied learning Koopman observables directly from data using neural networks.
\citet{takeishiLearningKoopmanInvariant2017} and \citet{yeungLearningDeepNeural2019} both learn nonlinear lifting functions and then fit linear dynamics in the lifted space. The first is more explicitly motivated by learning invariant subspaces for DMD and spectral analysis, while the second is framed as deep dictionary learning to avoid manually designing EDMD observables. Related deep operator learning approaches have also been proposed in molecular kinetics, including time-lagged autoencoders and VAMPnets \citep{mardtVAMPnetsDeepLearning2018, wehmeyerTimelaggedAutoencodersDeep2018}. \citet{luschDeepLearningUniversal2018} developed parsimonious, interpretable neural representations of Koopman eigenfunctions by embedding the dynamics on a low-dimensional manifold using an autoencoder architecture. A central difficulty is constructing or learning suitable lifted observables in the presence of inputs, particularly while preserving causal structure~\citep{selbyLearningCausalObservable2021}. 
Recent neural system-identification approaches with inputs, such as Koopman-Subnet~\citep{iacobDeepIdentificationNonlinear2021} and CT-Subnet~\citep{beintemaContinuoustimeIdentificationDynamic2022}, show that structured latent models can be effective for nonlinear dynamics. Adaptive Koopman-style architectures such as MamKO~\citep{liMamKOMambabasedKoopman2024} use historical context to modulate latent operators, combining Koopman-inspired modeling with Mamba~\citep{guMambaLinearTimeSequence2024}. However, these methods are primarily motivated by system identification or control rather than multi-step forecasting with exogenous variables.
Our work builds on Koopman-with-input formulations by shifting the emphasis from lifting construction to forecasting architecture. Treating the lift as a learnable component lets us focus on two key design questions for exogenous forecasting: how to reconstruct a forecast-start latent state from past context, and how to roll that state forward under future-known exogenous drivers. This yields a structured forecasting architecture that turns Koopman-inspired latent modeling into a practical multi-step prediction design.

\textbf{Koopman-inspired forecasting and structured linear sequence models.}
Recent work has adapted the ideas from Koopman theory of learning a latent space in which nonlinear dynamics evolve approximately linearly to forecasting settings. 
Autoencoder architectures have been proposed to learn latent representations for Koopman-inspired sequential data forecasting~\citep{azencotForecastingSequentialData2020}.
Spectral approaches have likewise been explored for forecasting quasi-periodic dynamical systems~\citep{langeFourierKoopmanSpectral2021}. 
The Koopman Neural Forecaster~\citep{wangKoopmanNeuralOperator2022} is a Koopman-based architecture designed for time series that exhibit temporal distribution shifts. 
Koopa~\citep{liuKoopaLearningNonstationary2023} addresses nonstationarity by decomposing time series into time-varying and time-invariant components.
In parallel, renewed interest in linear recurrent dynamics provides a complementary perspective.
Structured linear RNN models can match or exceed more complex recurrent baselines on long-context problems~\citep{orvietoResurrectingRecurrentNeural2023}, while SKOLR~\citep{zhangSKOLRStructuredKoopman2025a} provided an explicit connection between a linear recurrent architecture and Koopman theory. 
More broadly, state-space models such as S4~\citep{guEfficientlyModelingLong2022} and Mamba~\citep{guMambaLinearTimeSequence2024} show that structured latent linear dynamics can serve as highly effective sequence backbones. Hybrid approaches such as MamKO~\citep{liMamKOMambabasedKoopman2024} further connect this state-space modeling perspective with Koopman-style latent operators. Unlike such general-purpose or control-oriented sequence models, \modelname{} is designed specifically for exogenous multi-step forecasting. In particular, the Koopman forecasting and structured sequence models considered above do not jointly address the setting in which heterogeneous future-known exogenous covariates explicitly drive the latent dynamics throughout the forecast horizon.

\end{document}